\documentclass[journal]{IEEEtran}

\usepackage{graphicx}
\usepackage{amsmath,amsthm}
\usepackage{flushend}
\usepackage{algorithm}
\usepackage{algorithmic}
\usepackage{url}

\usepackage{array}
\usepackage{multirow}
\usepackage{makecell}

\newcommand{\vcenterrow}[3]{\multirow[c]{#1}{=}[#2]{\centering\arraybackslash #3}}

\usepackage{amssymb}
\usepackage{bbding}
\usepackage[compress]{cite}
\usepackage{balance}
\usepackage{booktabs}
\usepackage{ltxtable}
\usepackage{fancyvrb}
\usepackage{cases}
\usepackage{indentfirst}
\usepackage{ifthen}
\usepackage{threeparttable}
\usepackage{setspace}
\usepackage{color}
\usepackage{amsfonts}
\usepackage{csvsimple}

\usepackage{tikz}
\usepackage{pgfplots}
\pgfplotsset{width=8cm,compat=1.13}
\usepackage{arydshln} 
\usetikzlibrary{arrows,shapes,shadows,positioning,automata,backgrounds}
\usetikzlibrary{decorations.pathmorphing}
\usetikzlibrary{decorations.markings}
\usetikzlibrary{decorations.pathreplacing}

\theoremstyle{definition}

\theoremstyle{remark}

\usepackage{ragged2e}

\begin{document}


\title{LLM-Enabled UAV Dispatch: A System-Level Survey and Taxonomy}

%
%

\author{Xiao Han, Aoyang Quan, Xiangyu Zhao, Xiangjie Kong,~\IEEEmembership{Senior Member,~IEEE}, Guojiang Shen
\thanks{Xiao Han, Aoyang Quan, Guojiang Shen and Xiangjie Kong are with the School of Computer Science, Zhejiang University of Technology, China (e-mail: hahahenha@gmail.com; 19941348551@163.com; xjkong@ieee.org; gjshen1975@zjut.edu.cn).}
}

%
%

\maketitle

\begin{abstract}
\justifying
Unmanned aerial vehicle (UAV) dispatch is beginning to move beyond isolated path planning and optimization-driven resource allocation toward system-level coordination supported by semantic reasoning and LLM-based interfaces.
This survey provides a unified characterization of LLM-enabled UAV dispatch systems that bridges semantic intent, symbolic decision-making, and physical UAV execution.
Rather than treating LLMs as standalone add-ons, we conceptualize them as a cross-layer semantic orchestration layer connecting human instructions, external solvers, and distributed control modules.
We organize the literature into four representative dispatch paradigms: pipeline dispatch, global assignment dispatch, decentralized agentic dispatch, and divide-and-conquer dispatch.
For each paradigm, we analyze its decision logic, system structure, control flow, representative methods, and potential LLM roles.
We further examine how LLMs support semantic parsing, retrieval-grounded planning, solver orchestration, local agent reasoning, multi-agent coordination, safety assessment, and human-facing explanation.
We discuss the implications of these paradigms for scalability, robustness, coordination burden, and verification requirements, and identify open challenges including latency-aware reasoning, grounding reliability, physical feasibility guarantees, edge deployment, privacy protection, and distributed consistency.
This survey provides a system-level taxonomy and design perspective for integrating LLMs into safety-critical UAV dispatch systems.
\end{abstract}


\section{Introduction}
\label{sec:introduction}

\begin{figure*}[!t]
    \centering
    \includegraphics[width=\textwidth]{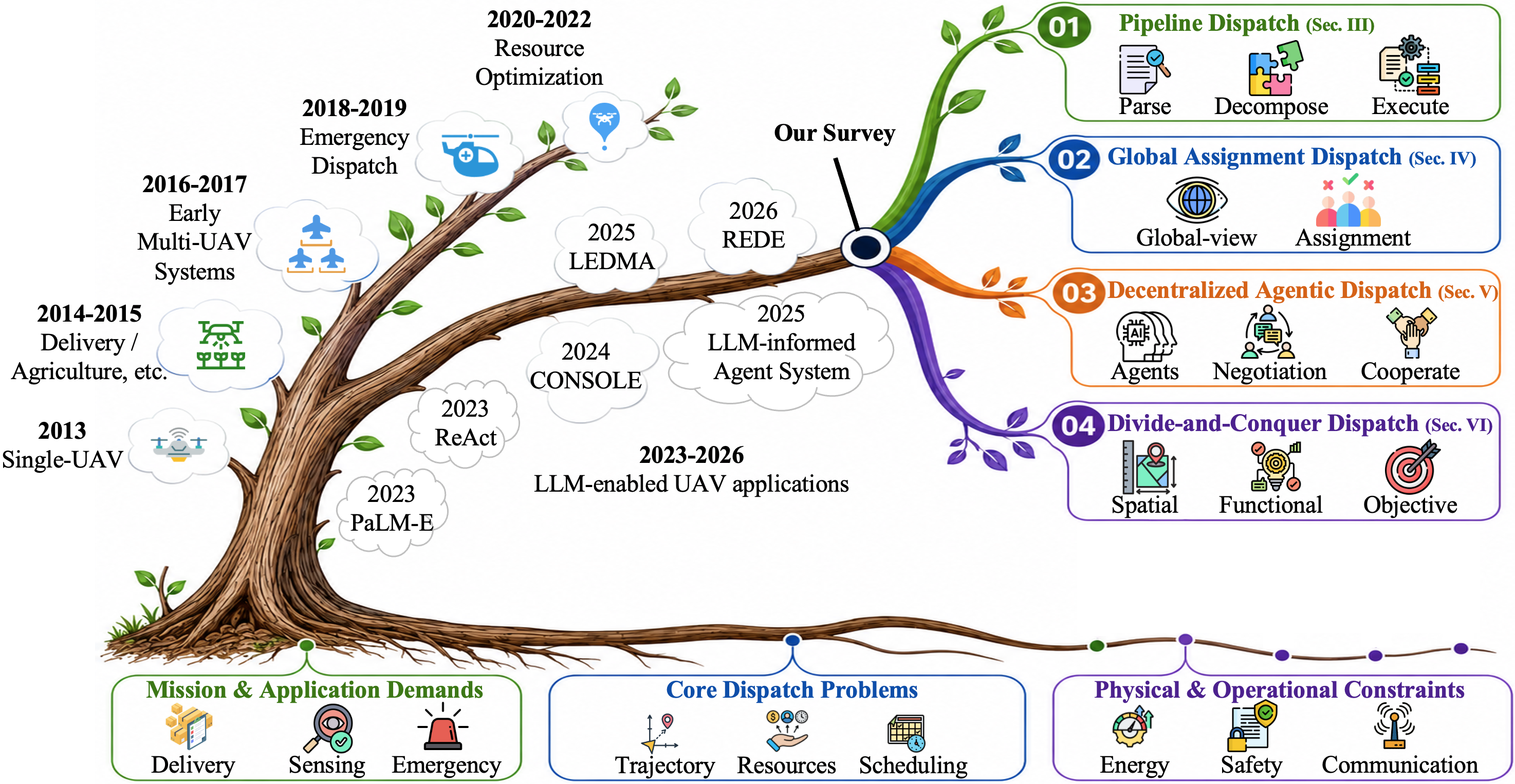}
    \caption{Evolution and scope of LLM-enabled UAV dispatch.}
    \label{fig:intro_overview}
\end{figure*}

\IEEEPARstart{U}{nmanned} Aerial Vehicles (UAVs) have evolved from small remote-controlled platforms into core infrastructure for the low-altitude economy (LAE) \cite{Pongsakornsathien_2025}.
They are now used in last-mile package delivery~\cite{11426817}, environmental monitoring and surveillance \cite{tagle2025effective}, disaster-zone emergency response \cite{manzini2026deploying}, infrastructure inspection \cite{ren2025}, and aerial communication services.
In these applications, the critical challenge has shifted from single-UAV navigation to fleet-level dispatch and coordination.
The dispatch system must decide which UAV should serve which task, when it should depart, which route it should follow, how energy and communication resources should be allocated, and how conflicts in dense low-altitude airspace should be avoided.
We refer to this system-level problem as \textbf{UAV dispatch}: the scheduling, routing, resource allocation, association, and coordination of UAV fleets under physical, communication, safety, privacy, and regulatory constraints.
Fig.~\ref{fig:intro_overview} organizes it as a tree: its roots ground UAV dispatch in mission and application demands, core dispatch problems, and physical and operational constraints.
Its trunk traces the evolution from single-UAV systems to LLM-enabled applications, and its four branches show the paradigms we summarized in this survey.

UAV dispatch is attracting rapidly growing attention in both research and real-world deployment.
Commercial UAV markets are expanding in logistics, inspection, and public services, regulators such as the FAA and EASA are developing frameworks for high-density BVLOS and U-space operations,
and China has positioned the LAE as a national growth engine.
This deployment momentum motivates the central premise of this survey: large-scale UAV operations cannot be supported by isolated path planning or single-agent control alone. 

As the number of UAVs grows, collision-free trajectories and communication resources become harder to allocate jointly \cite{Gong_2026}.
Early multi-quadrotor swarm systems \cite{6577992} further show that scalable UAV coordination requires tight integration among planning, communication, localization, and collision avoidance.
Dispatch decisions become tightly coupled across task assignment, trajectory planning, communication, energy, timing, and privacy.
Human operators increasingly issue open-ended mission requests rather than fully structured optimization inputs.
UAV dispatch therefore requires a framework that connects classical optimization and learning with semantic mission understanding, cross-layer coordination, and trustworthy human oversight.

The emergence of LLMs makes this framework more tractable by providing new interfaces for semantic interpretation, tool use, and human-facing explanation.
Recent vision-language and multimodal LLM studies provide the perception and grounding foundations needed for UAV missions involving aerial images, maps, and natural-language instructions \cite{han2023vitSurveyTPAMI,zhang2024vlmSurveyTPAMI,xu2025lvlmEhubTPAMI,yang2026defyingDistractionsTPAMI,huang2025causalityMLLMTPAMI,zhang2026beyondLlavaHdTPAMI,wang2026vlBiasBenchTPAMI,radford2021clip,li2023blip2,liu2023llava}.
Together with tool-use, agent, and embodied-model frameworks, these models can interpret natural-language missions, retrieve regulations or maps, reason over heterogeneous context, call external solvers, critique candidate plans, and explain dispatch decisions to human operators \cite{Chang_2024,shen2023hugginggpt,schick2023toolformer,yao2023react,besta2024graphOfThoughts,driess2023palmE}.

Recent UAV-oriented studies have already demonstrated a range of LLM-assisted capabilities, including path planning, vision-language navigation grounding, compliance-aware trajectory generation, multi-objective UAV network optimization, and edge-based multi-LLM orchestration \cite{Gong_2026,lin2024correctableLandmarkTPAMI,wang2026imagineNavTPAMI,lin2026evolveNavTPAMI,chen2025constraintAwareVLNTPAMI,li2025llmIsacUav,luo2025multiLlmEdge,liu2025llmInspection}.
However, these efforts remain fragmented across planning, communication, swarm coordination, inspection, and safety, lacking a unified abstraction to connect them.
Existing surveys summarized in Table~\ref{tab:intro_survey_comparison} predominantly focus on UAV trajectory prediction and path planning \cite{shukla2025,ghambari2024}, swarm networking and collaborative control \cite{cao2024,tang2023}, security and privacy \cite{sciancalepore2024}, or broader LLM-related topics such as model evaluation, vision-language models, and LLM-agent systems \cite{Chang_2024,zhang2024vlmSurveyTPAMI,guo2024llmMultiAgentsSurvey}.
However, these surveys still lack a dispatch-centered taxonomy that systematically explains how LLMs reshape the end-to-end UAV operation pipeline—from mission intent and task assignment to trajectory planning, scheduling, resource allocation, and execution monitoring.
In contrast, this work takes UAV dispatch as the central organizing abstraction, rather than using path planning, swarm control, or LLM agents as isolated lenses.
We study how LLMs reshape dispatch interfaces, dispatch logic, and dispatch verification across four emerging paradigms, thereby filling this gap and motivating the central question of this paper: \textbf{How should UAV dispatch be organized in the LLM era?}

\begin{table*}[t]
\centering
\caption{Comparison with representative surveys related to UAV dispatch and LLM-enabled aerial intelligence.}
\label{tab:intro_survey_comparison}
\begingroup
\scriptsize
\setlength{\arrayrulewidth}{0.4pt}
\setlength{\tabcolsep}{5pt}
\renewcommand{\arraystretch}{1.18}
\begin{tabular}{|>{\centering\arraybackslash}m{0.08\textwidth}
                |>{\arraybackslash}m{0.38\textwidth}
                |>{\arraybackslash}m{0.48\textwidth}|}
\hline
\textbf{Surveys} & \textbf{Focus \& Scope} & \textbf{Key Findings / Application Categories} \\
\hline
\cite{Pongsakornsathien_2025} &
Reviews low-altitude airspace management for uncrewed aircraft and advanced air mobility. &
Summarizes airspace concepts, operational requirements, BVLOS integration, regulatory constraints, and traffic-management challenges. \\
\hline

\cite{shukla2025} &
Surveys trajectory-prediction techniques for UAVs across modeling, data-driven learning, and motion-pattern analysis. &
Classifies trajectory-prediction methods and discusses their role in collision avoidance, motion forecasting, and safe UAV operation. \\
\hline

\cite{ghambari2024} &
Reviews UAV path-planning techniques with emphasis on planning algorithms, environment representation, and motion constraints. &
Categorizes path-planning approaches by optimization strategy, search mechanism, and constraint handling, with limited attention to end-to-end dispatch logic. \\
\hline

\cite{cao2024} &
Surveys computational-intelligence algorithms for UAV swarm networking and collaboration. &
Reviews communication, cooperative control, formation, search, learning-based collaboration, and future swarm-networking directions. \\
\hline

\cite{tang2023} &
Reviews swarm-intelligence algorithms for multi-UAV collaboration. &
Summarizes distributed cooperation, formation, task coordination, search, and collective decision-making mechanisms for UAV swarms. \\
\hline

\cite{sciancalepore2024} &
Reviews privacy and confidentiality issues in drone operations. &
Identifies privacy leakage, data exposure, trust, authentication, and secure communication challenges in UAV deployment. \\
\hline

\cite{Chang_2024} &
Surveys evaluation methods for large language models. &
Reviews evaluation dimensions for knowledge, reasoning, alignment, robustness, safety, and human preference, without targeting UAV dispatch. \\
\hline

\cite{zhang2024vlmSurveyTPAMI} &
Surveys vision-language models for vision tasks. &
Summarizes architectures, pretraining objectives, visual grounding, downstream vision-language tasks, and evaluation trends relevant to perception-rich UAV missions. \\
\hline

\cite{guo2024llmMultiAgentsSurvey} &
Surveys LLM-based multi-agent systems and their progress and challenges. &
Reviews agent profiling, memory, planning, tool use, communication, cooperation, and benchmarks, but does not address physical UAV dispatch constraints. \\
\hline

\textbf{This survey} &
UAV dispatch in the LLM era, centered on the transformation from mission intent to executable fleet decisions. &
Organizes LLM-enabled UAV operation into pipeline dispatch, global assignment dispatch, decentralized agentic dispatch, and divide-and-conquer dispatch, with LLM roles as parsers, planners, solver orchestrators, local agents, coordinators, critics, and interfaces. \\
\hline
\end{tabular}
\endgroup
\end{table*}

To answer this question, we propose a dispatch-logic taxonomy based on the transformation from mission descriptions to executable dispatch decisions.
At the top level, we organize dispatch logic into four paradigms: pipeline dispatch, global assignment dispatch, decentralized agentic dispatch, and divide-and-conquer dispatch.
After this top-level classification, we organize papers within each paradigm along several secondary axes.
These axes include the dispatch decisions they cover, namely task assignment, trajectory or route planning, scheduling, resource allocation, and cooperation or association among UAVs.
We also record the solver family used by each study, including optimization, learning, auctions, heuristics, or LLM-tool orchestration, and the application context it targets.

Among these categories, \textbf{(1) pipeline dispatch} covers methods that convert a mission into ordered stages, such as semantic parsing, task decomposition, assignment, trajectory generation, scheduling, and monitoring.
Its dispatch process is organized as a stage-wise workflow in which LLMs can serve as parsers, tool selectors, and critics, rather than as a single global optimization problem.
\textbf{(2) Global assignment dispatch} covers centralized or globally coordinated methods that jointly determine task matching, routes, schedules, resources, and associations when sufficient global state is available.
It relies on a central dispatcher or equivalent global view, while the LLM layer interprets intent, retrieves rules, explains tradeoffs, or guides solver choice.
\textbf{(3) Decentralized agentic dispatch} covers agent-led methods in which UAVs locally select tasks, bid, negotiate, replan, or learn policies from partial observations.
Building on this decentralized decision-making paradigm, task ownership emerges from local autonomy, while LLM agents serve as complementary components that enhance the system with memory, semantic negotiation, and reflection, without replacing verified control mechanisms.
\textbf{(4) Divide-and-conquer dispatch} covers methods that partition a large mission by dispatch decisions, objectives, regions, time windows, fleet groups, roles, communication links, knowledge sources, or model placement.
The main design question is how to split and then reconnect subproblems through solvers, retrieval sources, edge models, and multi-agent orchestration.
Together, these four paradigms cover the dispatch layer of UAV systems.

The main contributions of this paper are as follows:
\begin{itemize}
    \item We define UAV dispatch in the LLM era as a
    semantic--symbolic--physical decision process that maps human
    intent and environmental context to task assignment, trajectory,
    schedule, resource allocation, and cooperation decisions.
    
    \item We propose a dispatch-logic taxonomy with four LLM-era paradigms.
    This taxonomy connects classical UAV dispatch methods with emerging LLM-enabled dispatch architectures.
    
    
    \item We analyze how LLMs can serve as semantic parsers,
    retrieval-grounded planners, solver orchestrators, local agents, multi-agent coordinators, safety critics, and human-facing explanation interfaces, while emphasizing that physical feasibility must still be guaranteed by specialized modules.
    
    \item We identify open challenges and future directions for
    trustworthy LLM-enabled UAV dispatch, including grounding,
    hallucination control, edge deployment, multi-agent trust,
    privacy, accountability, benchmarking, and human governance.
\end{itemize}

The remainder of this paper is organized as follows.
Section~\ref{sec:preliminaries} introduces the preliminaries and
unified notation for UAV dispatch. Sections~\ref{sec:pipeline_dispatch} to~\ref{sec:divide_conquer_dispatch} review the four dispatch
paradigms: pipeline dispatch for UAV missions, global assignment
dispatch, decentralized agentic dispatch, and divide-and-conquer dispatch.
Section~\ref{sec:opportunities_challenges} synthesizes the four paradigms in the LLM era and discusses future challenges and research
directions. Section~\ref{sec:conclusion} concludes the survey.

\section{Preliminaries}
\label{sec:preliminaries}

Before reviewing the four paradigms in detail, we first introduce the definitions and notation used throughout this survey.
We begin with the traditional dispatch framework, which provides the classical notions of task assignment, scheduling, routing, and resource allocation.
We then extend this view to unmanned dispatch systems, emphasizing how autonomous perception, communication, and online adaptation inherit from, yet differ from, traditional dispatch.
Finally, we specialize this view to UAV dispatch systems, where UAV operations introduce three-dimensional mobility, airspace safety, endurance, communication, and regulatory constraints.

\subsection{Traditional Dispatch Framework}
\label{subsec:traditional_dispatch_framework}

Traditional dispatch studies how a set of service resources should be matched with a set of spatially or temporally distributed requests.
The agents may be vehicles, robots, workers, ships, or aircraft, while the requests may represent delivery orders, inspection targets, emergency events, passengers, or sensing tasks.
To keep the notation consistent with later sections, we use $\mathcal{U}=\{u_1,u_2,\ldots,u_N\}$ for the agent set, $\mathcal{T}=\{\tau_1,\tau_2,\ldots,\tau_M\}$ for the task set, and $\Omega$ for the operating environment.
The core decisions are assignment $\mathbf{X}$, routing or trajectory planning $\mathbf{P}$, scheduling $\mathbf{S}$, resource allocation $\mathbf{R}$, and association or cooperation $\mathbf{A}$.

The assignment matrix $\mathbf{X}=[x_{ij}]$ indicates whether agent $u_i$ serves task $\tau_j$, with $x_{ij}=1$ if the assignment is made and $x_{ij}=0$ otherwise.
A feasible traditional dispatch plan covers the required tasks, respects capacity, endurance, and service-time limits, and keeps the selected routes feasible in $\Omega$.
Such a formulation captures classical problems including vehicle routing, pickup-and-delivery, ride-hailing, warehouse robot coordination, emergency vehicle dispatch, and field-service scheduling.
Its key role is to determine \textbf{who} serves each task, \textbf{when} it is served, \textbf{where} the agent moves, and \textbf{how} limited resources are allocated.

\subsection{Unmanned Dispatch System}
\label{subsec:unmanned_dispatch_system}

Unmanned dispatch extends the traditional framework by replacing passive executors with autonomous agents.
Each agent may perceive the environment, communicate with neighbors, update its local state, and adapt its actions during execution. The state of the unmanned agent $u_i$
at time $t$ can be summarized compactly as $\mathbf{s}_i(t)=(\mathbf{q}_i(t),e_i(t),o_i(t))$, where $\mathbf{q}_i(t)$ is the position and $e_i(t)$ is the remaining energy.
The term $o_i(t)$ aggregates local observations such as nearby tasks, obstacles, communication status, capability mode, and map information.
The system state is then the collection of all agent states, task states, and environmental conditions.

Compared with traditional dispatch, unmanned dispatch has three additional features.
First, the environment is often partially observable, so each agent may have only local or delayed information.
Second, task execution and motion planning are tightly coupled because feasibility depends on current position, energy, sensing range, communication quality, and mobility limits.
Third, the dispatch logic must be online and adaptive, since new tasks, obstacles, failures, or communication interruptions may appear during execution.
Therefore, unmanned dispatch bridges classical operations research and autonomous multi-agent coordination.

\subsection{UAV Dispatch Methods}
\label{subsec:aerial_dispatch_methods}

UAV dispatch further specializes in unmanned dispatch to low-altitude airspace and airborne agents.
In this survey, it refers to the coordination of UAVs for delivery, inspection, sensing, surveillance, communication, emergency response, and aerial data collection.
Compared with ground robots or vehicles, UAVs operate in three-dimensional space
and must jointly satisfy flight dynamics, airspace safety, battery endurance, communication coverage, weather, regulation, and privacy constraints.

For UAV $u_i$, let $\mathbf{q}_i(t)=[x_i(t),y_i(t),z_i(t)]^\top$ denote its position at time $t$.
A feasible trajectory must satisfy mobility limits such as $\|\dot{\mathbf{q}}_i(t)\|\leq v_i^{\max}$, collision avoidance $\|\mathbf{q}_i(t)-\mathbf{q}_k(t)\|\geq d_{\min}$ for $i\neq k$, airspace restrictions in $\Omega$, endurance requirements expressed through $e_i(t)$, and communication-quality requirements for the assigned UAV--task pairs.
These constraints make UAV dispatch a coupled problem of assignment, trajectory planning, timing, communication, and resource management.

For the UAV dispatch problem considered in this survey, an LLM-enabled interface may transform human instructions and contextual information into structured tasks, constraints, and preferences, while the dispatch layer optimizes the resulting multi-objective planning
problem.
In this view, the LLM does not replace the optimization model.
Instead, it provides a semantic interface layer: retrieval-augmented generation (RAG) can ground the prompt in maps, regulations, weather reports, historical logs, and mission documents; structured reasoning prompts or rationale generation can make intermediate dispatch assumptions more inspectable; and self-refinement or external evaluators can check whether a candidate plan satisfies physical, regulatory, and privacy constraints \cite{lewis2020rag,wei2022chainOfThought,schick2023toolformer,yao2023treeOfThoughts,besta2024graphOfThoughts,madaan2023selfRefine,romera2024funsearch}.
We use the following objective as a unified abstraction:
\begin{equation}
\begin{aligned}
    \min_{\mathbf{X},\mathbf{P},\mathbf{S},\mathbf{R},\mathbf{A}}
    \quad
    J(&\mathbf{X},\mathbf{P},\mathbf{S},\mathbf{R},\mathbf{A})
    =
    \lambda_1 C_{\mathrm{time}}(\mathbf{X},\mathbf{P},\mathbf{S}) \\
    &+ \lambda_2 C_{\mathrm{energy}}(\mathbf{X},\mathbf{P},\mathbf{R})
    + \lambda_3 C_{\mathrm{delay}}(\mathbf{X},\mathbf{S}) \\
    &+ \lambda_4 C_{\mathrm{risk}}(\mathbf{P},\mathbf{A};\Omega)
    + \lambda_5 C_{\mathrm{comm}}(\mathbf{P},\mathbf{R},\mathbf{A}) \\
    &+ \lambda_6 L_{\mathrm{priv}}(\mathbf{X},\mathbf{R},\mathbf{A}) \\
    &- \lambda_7 Q_{\mathrm{serv}}(\mathbf{X},\mathbf{P},\mathbf{S},\mathbf{R},\mathbf{A}),
\end{aligned}
\label{eq:general_objective}
\end{equation}
where $\mathbf{X}$, $\mathbf{P}$, $\mathbf{S}$, $\mathbf{R}$, and $\mathbf{A}$ denote assignment, trajectory or path, schedule, resource allocation, and association or cooperation decisions, respectively.
$C_{\mathrm{time}}(\mathbf{X},\mathbf{P},\mathbf{S})$ measures mission time, $C_{\mathrm{energy}}(\mathbf{X},\mathbf{P},\mathbf{R})$ measures energy consumption, and
$C_{\mathrm{delay}}(\mathbf{X},\mathbf{S})$ measures service delay.
$C_{\mathrm{risk}}(\mathbf{P},\mathbf{A};\Omega)$ measures safety risk under the operating environment $\Omega$, $C_{\mathrm{comm}}(\mathbf{P},\mathbf{R},\mathbf{A})$ measures
communication cost, and $L_{\mathrm{priv}}(\mathbf{X},\mathbf{R},\mathbf{A})$ measures privacy leakage.
$Q_{\mathrm{serv}}(\mathbf{X},\mathbf{P},\mathbf{S},\mathbf{R},\mathbf{A})$ measures service quality such as coverage, throughput, sensing accuracy, or task completion ratio.
The non-negative weights $\lambda_1,\ldots,\lambda_7$ balance these objectives for different
UAV dispatch scenarios.

\section{Pipeline Dispatch}
\label{sec:pipeline_dispatch}

\begin{figure}[t]
    \centering
    \includegraphics[width=0.85\columnwidth]{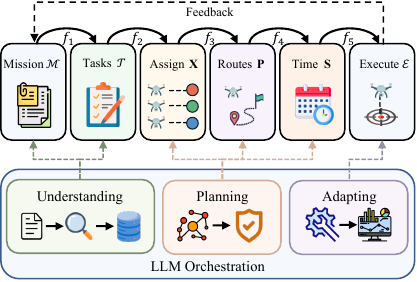}
    \caption{Pipeline dispatch decomposes a mission into ordered processing stages, where an LLM layer can parse intent, call tools, and trigger verification.}
    \label{fig:pipeline_dispatch}
\end{figure}

Pipeline dispatch for UAV missions decomposes a complex mission into ordered modules, so that semantic parsing, task decomposition, assignment, trajectory generation, scheduling, and execution monitoring are handled as successive stages rather than as a single monolithic optimization problem.
Formally, given a mission description $\mathcal{M}$, a pipeline-dispatch framework can be abstracted as:
\begin{equation}
    \mathcal{M}
    \xrightarrow{f_1}
    \mathcal{T}
    \xrightarrow{f_2}
    \mathbf{X}
    \xrightarrow{f_3}
    \mathbf{P}
    \xrightarrow{f_4}
    \mathbf{S}
    \xrightarrow{f_5}
    \mathcal{E},
\label{eq:pipeline_framework}
\end{equation}
where $\mathcal{M}$ denotes the mission description and $\mathcal{E}$ denotes the execution or monitoring stage.
The functions $f_1,\ldots,f_5$ may correspond to task decomposition, clustering, assignment, path planning, scheduling, collision avoidance, or online replanning, depending on the application scenario.
In the LLM era, the same abstraction becomes a tool-grounded reasoning chain:
an LLM or multimodal LLM parses $\mathcal{M}$, retrieves
mission knowledge, decomposes the request into structured tasks, selects downstream solvers, and asks external validators to check the produced assignment, trajectory, schedule, and execution feedback \cite{lewis2020rag,shen2023hugginggpt,yao2023react,Gong_2026}.
Thus, the LLM layer does not replace the pipeline; it makes the front end more adaptive to natural-language intent, heterogeneous context, and changing low-altitude constraints while keeping each stage auditable.
Figure~\ref{fig:pipeline_dispatch} illustrates this stage-wise dispatch logic.

\begin{table*}[t]
\centering
\caption{Taxonomy of representative pipeline dispatch methods and their LLM-era orchestration roles aligned with Eq.~\eqref{eq:pipeline_framework}.} \label{tab:pipeline_subtask_coverage} \footnotesize \setlength{\arrayrulewidth}{0.4pt} \setlength{\tabcolsep}{3pt} \renewcommand{\arraystretch}{1.08} \begin{tabular}{|>{\raggedright\arraybackslash}m{0.20\textwidth} |>{\raggedright\arraybackslash}m{0.19\textwidth} |>{\raggedright\arraybackslash}m{0.24\textwidth} |>{\raggedright\arraybackslash}m{0.30\textwidth}|} \hline \textbf{Pipeline Transition} & \textbf{Classical Stage Role} & \textbf{LLM-Era Orchestration Role} & \textbf{Representative Methods} \\ \hline \vcenterrow{3}{-3ex}{$f_1:\mathcal{M}\!\rightarrow\!\mathcal{T}$} & Spatial or task decomposition & Semantic scene parsing and RAG-grounded task graph construction & Li \textit{et al.}~\cite{li2025hierarchical}, Xiong \textit{et al.}~\cite{xiong2025} \\ \cline{2-4} & Hierarchical mission decomposition & Natural-language mission breakdown and role prompting & Luan \textit{et al.}~\cite{luan2022hierarchical}, Cheng \textit{et al.}~\cite{2025Mission}, Guo \textit{et al.}~\cite{guo2025hierarchical}, Li \textit{et al.}~\cite{li2025hierarchical}, J\"{u}nger \textit{et al.}~\cite{junger2021talking} \\ \cline{2-4} & Rolling or adaptive clustering & Context-aware reclustering from execution feedback & Yanmaz \textit{et al.}~\cite{yanmaz2024}, Ren \textit{et al.}~\cite{ren2025} \\ \hline \vcenterrow{3}{-2ex}{$f_2:\mathcal{T}\!\rightarrow\!\mathbf{X}$} & Task or role assignment & Tool-call handoff from task graph to assignment solver & Yanmaz \textit{et al.}~\cite{yanmaz2024}, Guo \textit{et al.}~\cite{guo2025hierarchical}, Zhao \textit{et al.}~\cite{zhao2026multi}, Li \textit{et al.}~\cite{li2025hierarchical}, Gong \textit{et al.}~\cite{Gong_2026} \\ \cline{2-4} & Hierarchical task allocation & Constraint generation and priority explanation & Luan \textit{et al.}~\cite{luan2022hierarchical}, Cheng \textit{et al.}~\cite{2025Mission}, Guo \textit{et al.}~\cite{guo2025hierarchical} \\ \cline{2-4} & Battery or inspection-task matching & Resource-state summarization and repair hints & Alyassi \textit{et al.}~\cite{alyassi2023}, Ren \textit{et al.}~\cite{ren2025} \\ \hline \vcenterrow{4}{-6ex}{$f_3:\mathbf{X}\!\rightarrow\!\mathbf{P}$} & Communication-aware trajectory design & Constraint interpretation before path optimization & Wang \textit{et al.}~\cite{wang2022}, Jing \textit{et al.}~\cite{jing2024}, Zhang \textit{et al.}~\cite{zhang2024isac}, Li \textit{et al.}~\cite{li2025llmIsacUav} \\ \cline{2-4} & Learning-based trajectory control & Compliance-aware critic for learned policies & Zhu \textit{et al.}~\cite{zhu2024}, Gong \textit{et al.}~\cite{Gong_2026} \\ \cline{2-4} & Coverage, search, or sensing paths & Vision-language grounding of regions and targets & Yanmaz \textit{et al.}~\cite{yanmaz2024}, Li \textit{et al.}~\cite{li2025hierarchical}, Xiong \textit{et al.}~\cite{xiong2025}, Liu \textit{et al.}~\cite{liu2023aerialvln}, Lin \textit{et al.}~\cite{lin2024correctableLandmarkTPAMI,lin2025navcotTPAMI,lin2026evolveNavTPAMI}, Wu \textit{et al.}~\cite{wu2025aeroduo} \\ \cline{2-4} & Hierarchical or collision-aware path planning & External validator for safe path refinement & Cheng \textit{et al.}~\cite{2025Mission}, Zhou \textit{et al.}~\cite{zhou2025hierarchical} \\ \hline \vcenterrow{3}{-2ex}{$f_4:\mathbf{P}\!\rightarrow\!\mathbf{S}$} & Time allocation or service ordering & Schedule explanation and local repair & Wang \textit{et al.}~\cite{wang2022}, Jing \textit{et al.}~\cite{jing2024}, Zhang \textit{et al.}~\cite{zhang2024isac} \\ \cline{2-4} & Subtask or execution scheduling & Agent memory for rolling execution context & Luan \textit{et al.}~\cite{luan2022hierarchical}, Cheng \textit{et al.}~\cite{2025Mission}, Ren \textit{et al.}~\cite{ren2025} \\ \cline{2-4} & Battery and charging scheduling & Energy-state explanation and charging trigger & Alyassi \textit{et al.}~\cite{alyassi2023} \\ \hline \vcenterrow{3}{-2ex}{$f_5:\mathbf{S}\!\rightarrow\!\mathcal{E}$} & Online execution update & Reflection loop and replanning trigger & Zhu \textit{et al.}~\cite{zhu2024}, Guo \textit{et al.}~\cite{guo2025hierarchical}, Ren \textit{et al.}~\cite{ren2025} \\ \cline{2-4} & Connectivity maintenance and replanning & Communication-aware agent monitor & Yanmaz \textit{et al.}~\cite{yanmaz2024} \\ \cline{2-4} & Collision or safety handling & Runtime safety critic and human alert & Zhou \textit{et al.}~\cite{zhou2025hierarchical} \\ \hline \end{tabular} \end{table*}

\subsection{Hierarchical Task Decomposition and Scheduling}
\label{subsec:hierarchical_pipeline}

The first major line of pipeline dispatch emphasizes the decomposition architecture itself.
Instead of focusing only on the final trajectory, these methods study how a complex mission can be broken into subtasks, organized into multiple layers, and scheduled across UAVs in a structured manner.
%
This subsection corresponds mainly to the $f_1$, $f_2$, and $f_4$ rows of Table~\ref{tab:pipeline_subtask_coverage}: high-level missions are first decomposed, then assigned or allocated, and finally organized into executable schedules.

The basic hierarchical pattern is to separate mission understanding from execution ordering.
Luan \textit{et. al.} ~\cite{luan2022hierarchical} use an upper layer for topology reconstruction and task allocation and a lower layer for subtask scheduling in UAV-assisted mobile edge computing emergency networks.
Cheng \textit{et. al.}~\cite{2025Mission} extend the pattern to heterogeneous aircraft-carrier--UAV swarms, separating the carrier path planning layer from the multi-UAV collaborative task planning layer, while Guo \textit{et. al.}~\cite{guo2025hierarchical} use a hierarchical framework to connect task decomposition with marginal return allocation.
Across these studies, hierarchy is not only a computational trick; it defines which layer is responsible for mission abstraction, mobility support, and executable task ordering.

Once the hierarchy is established, the next challenge is how to refine allocation decisions without losing scalability.
Guo \textit{et. al.} ~\cite{guo2025hierarchical} cluster and decompose tasks in the upper layer, then allocate subtasks through marginal return optimization.
Zhao \textit{et. al.}~\cite{zhao2026multi} use a coarse initialization stage followed by a parallel auction network for multi-target tracking, turning global allocation into a staged refinement process.
Li \textit{et. al.}~\cite{li2025hierarchical} add task priority and multi-objective metaheuristic optimization to multi-region coverage.
These methods move beyond a fixed two-layer pipeline: decomposition produces an initial structure, while optimization inside later stages adapts the assignment to heterogeneity, priority, and scale.

Hierarchical pipelines are also useful when the input or the internal planning module is complex.
In autonomous exploration, hierarchical frontier-based planners such as FUEL show how global exploration objectives, frontier selection, and local trajectory generation can be organized into a fast hierarchical pipeline \cite{zhou2021fuel}.
J\"{u}nger \textit{et. al.} ~\cite{junger2021talking} bridge high-level human commands and low-level UAV subtasks through hierarchical task decomposition, which is directly relevant to LLM-enabled dispatch interfaces.
In such an interface, the upper layer should not be viewed only as a manually written task parser.
It can be an LLM-based semantic layer that retrieves mission rules, converts operator intent into a task graph, selects specialized perception or optimization tools, and records why
each subtask is created \cite{lewis2020rag,shen2023hugginggpt,yao2023react}.
Embodied multimodal models further show how language, visual observations, and continuous robot states can be grounded in one planning interface \cite{driess2023palmE}, while AerialVLN and large-model landmark discovery show that navigation requires spatial reasoning beyond ground-agent VLN \cite{liu2023aerialvln,lin2024correctableLandmarkTPAMI,wang2026imagineNavTPAMI,lin2026evolveNavTPAMI,lin2025navcotTPAMI}.
Zhou \textit{et. al.} ~\cite{zhou2025hierarchical} apply a related decomposition principle inside path planning: an upper layer generates initial Dubins-curve paths, while a lower layer resolves conflicts with a three-dimensional velocity-obstacle strategy.
These examples show that hierarchy can operate at different semantic levels, from natural-language command interpretation and evidence-grounded task construction to collision-aware trajectory refinement.

These hierarchical methods demonstrate that pipeline dispatch is not merely a simple engineering workflow.
It can serve as a general decomposition principle for organizing complex UAV missions, especially when the original problem contains multiple coupled decision layers.

\subsection{Trajectory-Centric Pipeline Dispatch}

This subsection classifies trajectory-centric studies as pipeline dispatch when trajectory generation is formulated as a downstream stage following task allocation or service-order decisions, regardless of whether the trajectory solver is centralized, distributed, optimization-based, or learning-based.
Within this family, trajectory generation constitutes the primary technical bottleneck.
The main focus is therefore on designing and optimizing UAV trajectories for communication, data collection, coverage, and target-tracking scenarios.
In the notation of Table~\ref{tab:pipeline_subtask_coverage}, these studies are centered on $f_3:\mathbf{X}\rightarrow\mathbf{P}$, while additional requirements such as service ordering, connectivity maintenance, and collision avoidance may further involve $f_4$ and $f_5$.

One trajectory-centric branch couples UAV movement with communication, sensing, and service objectives.
Wang \textit{et. al.}~\cite{wang2022} expose the upstream and downstream dependencies of this trajectory module: a service order is first selected, and the resulting path is optimized together with sensor uploading powers.
Recent ISAC-oriented studies further expand this coupling by coordinating trajectory, association, beamforming, and target-localization quality \cite{jing2024,zhang2024isac}, while LLM-guided UAV-ISAC optimization shows how language reasoning can guide multi-objective search before a numerical solver verifies feasibility \cite{li2025llmIsacUav}.
Together, these studies show that communication-aware trajectories are rarely standalone geometric paths; they usually inherit task-ordering, sensing, and link-quality requirements from adjacent pipeline stages.

When the environment is dynamic, the trajectory stage must also become adaptive.
Zhu \textit{et. al.}~\cite{zhu2024} introduce collaborative
reinforcement learning for three-dimensional target tracking, replacing a fixed offline trajectory optimizer with a learned motion policy that responds to changing target states.
Yanmaz \textit{et. al.}~\cite{yanmaz2024} extend this idea at the system level by separating search and relay roles: search UAVs follow targets, while relay UAVs preserve network connectivity.
These studies indicate that $f_3$ and $f_5$ become coupled when sensing, motion, and communication must be updated during the mission.

Coverage missions reveal another dependency: trajectory generation often requires an earlier spatial decomposition stage.
Recent multi-region and post-disaster studies make the same dependency more explicit by combining region partitioning, task allocation, and local coverage routing \cite{li2025hierarchical,xiong2025}.
Thus, even in trajectory-centric dispatch, the quality of $\mathbf{P}$ depends strongly on how the mission is decomposed into spatial or functional units.

In the LLM era, the trajectory stage is better understood as a grounded tool call rather than a free-form path generated by language alone.
The LLM or multimodal LLM can translate an instruction such as ``inspect the blocked bridge and avoid dense residential blocks'' into semantic waypoints, forbidden regions, and task priorities.
However, the final path still has to be produced or verified by geometry, control, and safety modules.
PaLM-E supports this view by connecting language with embodied sensory states \cite{driess2023palmE}, while
vision-language pretraining, open-vocabulary perception, and segmentation foundation models provide the perception substrate for turning aerial imagery into structured targets or regions \cite{han2023vitSurveyTPAMI,zhang2024vlmSurveyTPAMI,zhu2024openVocabularySurveyTPAMI,zhu2025llafsTPAMI,radford2021clip,alayrac2022flamingo,li2023blip2,kirillov2023segmentAnything}.
AerialVLN-style benchmarks make the aerial grounding problem explicit for UAVs \cite{liu2023aerialvln,wu2025aeroduo}.
Recent hybrid DRL--LLM trajectory planning further suggests that LLM reasoning is most useful as a compliance-aware critic and constraint interpreter, while the low-level policy handles motion feasibility \cite{Gong_2026}.

Overall, trajectory-centric pipeline dispatch treats task decomposition as a lightweight preprocessing step and places the main optimization burden on path or trajectory generation.
Such methods are effective when the task structure is relatively clear, and the key bottleneck is mobility optimization.
However, they may become less effective when task decomposition, resource management, and trajectory planning are strongly coupled.

\subsection{Resource-Aware and Application-Oriented Pipelines}
\label{subsec:resource_aware_pipeline}

Beyond trajectory design and task decomposition, the pipeline paradigm has also been applied to resource management and application-specific dispatch.
In these studies, the key idea is to isolate a critical
resource or operational factor as an independent stage of the dispatch process.
Compared with the previous two subsections, the organizing logic here is less about a universal decomposition architecture and more about which bottleneck should be promoted to a first-class pipeline stage.

Inspection scenarios highlight the need for rolling updates across $f_1$, $f_2$, $f_4$, and $f_5$.
Application-oriented pipelines must couple visual target tracking, trajectory generation, and control feasibility during execution, as in autonomous power-line inspection with perception-aware MPC \cite{xing2023autonomous}.
Ren \textit{et. al.}~\cite{ren2025} first cluster inspection tasks according to task characteristics and then apply greedy partition scheduling to imitate manual planning logic.
Because long-duration inspection missions may receive updated task sets during execution, the pipeline is repeatedly refreshed rather than solved once.
This makes rolling clustering and scheduling the core mechanism, while trajectory planning becomes a downstream consequence of the current inspection workload.

Delivery and long-duration inspection scenarios expose a different bottleneck: energy availability.
Alyassi \textit{et. al.}~\cite{alyassi2023} jointly plan autonomous recharging and flight missions for battery-operated drones.
Here, the pipeline is organized around energy logistics rather than only around UAV movement; feasibility depends on whether the right energy state can be assigned at the right time.


In LLM-enabled application pipelines, mission knowledge becomes another dispatch resource.
Inspection reports, emergency messages, weather alerts, battery logs, and airspace notices may arrive as heterogeneous text, images, or tables.
An LLM-informed drone inspection system can convert such inputs into candidate defects, inspection subtasks, and
agent responsibilities \cite{liu2025llmInspection}; self-refinement and reflection mechanisms can then turn execution feedback into repair suggestions without changing the verified scheduler or controller \cite{madaan2023selfRefine,shinn2023reflexion}.
Therefore, the application-oriented pipeline becomes a closed loop: retrieve domain evidence, extract constraints, call a solver, verify execution, and update the next dispatch prompt from monitored outcomes.
Aerial construction and manufacturing tasks provide another example of such closed-loop dispatch, where multiple aerial robots must coordinate task execution, geometric feedback, online correction, and human supervision during physical operation \cite{zhang2022aerial}.

These application-oriented works reveal that pipeline design is often driven by the dominant bottleneck of the target scenario.
Inspection, delivery, and emergency sensing pipelines differ mainly in their dominant operational constraint: rolling task updates, energy logistics, and range-aware partitioning, respectively.
Therefore, the structure of a pipeline is not fixed; it should be adapted to the mission objective, resource limitations, and operational environment.

\section{Global Assignment Dispatch}
\label{sec:assignment_dispatch}

\begin{figure}[t]
    \centering
    \includegraphics[width=0.85\columnwidth]{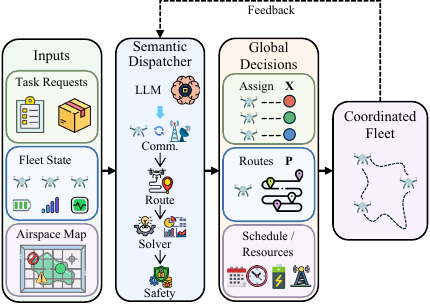}
    \caption{Global assignment dispatch uses a central dispatcher to jointly determine tasks, routes, schedules, and resources, with an LLM layer supporting intent interpretation and solver selection.}
    \label{fig:assignment_dispatch}
\end{figure}

Global assignment dispatch uses a central dispatcher, such as a control station, cloud platform, edge server, fleet manager, or LLM-assisted meta-dispatcher, to collect global information and jointly determine the main dispatch decisions introduced in Section~\ref{sec:preliminaries}.
Unlike pipeline dispatch, global assignment dispatch does not pass intermediate artifacts through a fixed sequence of stages. Instead, it jointly optimizes task matching $\mathbf{X}$, routes $\mathbf{P}$, schedules $\mathbf{S}$, resources $\mathbf{R}$, and associations $\mathbf{A}$ under a shared global state.
Classical multi-UAV coordination studies have already emphasized that goal assignment, resource allocation, and trajectory generation are tightly coupled in fleet-level control, which provides an early systems foundation for the global assignment view \cite{1242572}.
Table~\ref{tab:assignment_taxonomy} summarizes this paradigm according to four recurring dimensions: mobility--communication assignment, routing--scheduling assignment, scalable solver design, and security or privacy constraints.
In the LLM era, this global view can be wrapped by a semantic meta-dispatcher that translates operator intent into constraints, retrieves airspace or network knowledge, chooses assignment solvers, and explains trade-offs before the optimization core commits a plan.
Fig.~\ref{fig:assignment_dispatch} illustrates this global assignment view.

\begin{table*}[t]
\centering
\caption{Taxonomy of representative global assignment dispatch methods and their LLM-era assignment roles.}
\label{tab:assignment_taxonomy}
\footnotesize
\setlength{\arrayrulewidth}{0.4pt}
\setlength{\tabcolsep}{2pt}
\renewcommand{\arraystretch}{1.08}
\begin{tabular}{|>{\raggedright\arraybackslash}m{0.22\textwidth}
                |>{\raggedright\arraybackslash}m{0.22\textwidth}
                |>{\raggedright\arraybackslash}m{0.24\textwidth}
                |>{\raggedright\arraybackslash}m{0.28\textwidth}|}
\hline
\textbf{Central Dispatch Dimension} &
\textbf{Coupled Decision Role} &
\textbf{LLM-Era Assignment Role} &
\textbf{Representative Methods} \\
\hline
\vcenterrow{3}{-3ex}{Mobility--communication} &
Trajectory and link scheduling &
Global-state summarization and link-aware prompt grounding &
Liu and Zheng~\cite{liu2022}, He \textit{et al.}~\cite{he2023}, Hu \textit{et al.}~\cite{hu2024}, Jing \textit{et al.}~\cite{jing2024}, Zhang \textit{et al.}~\cite{zhang2024isac}, Wu \textit{et al.}~\cite{wu2023} \\
\cline{2-4}
&
Data collection, offloading, and MEC &
Service-intent parsing and resource constraint extraction &
Nguyen \textit{et al.}~\cite{nguyen2022}, Zhu \textit{et al.}~\cite{zhu2023}, Liu and Zheng~\cite{liu2022}, He \textit{et al.}~\cite{he2023}, Hu \textit{et al.}~\cite{hu2024}, Shi \textit{et al.}~\cite{shi2021} \\
\cline{2-4}
&
Association, wireless power, and ISAC &
Multi-objective tradeoff explanation and search guidance &
Jing \textit{et al.}~\cite{jing2024}, Zhang \textit{et al.}~\cite{zhang2024isac}, Wu \textit{et al.}~\cite{wu2023}, Li \textit{et al.}~\cite{li2025llmIsacUav} \\
\hline
\vcenterrow{3}{-2.5ex}{Routing--scheduling} &
TSP/VRP-style UAV routing &
Natural-language service request to routing constraints & Montemanni and Dell'Amico~\cite{montemannind}, Joo \textit{et al.}~\cite{joo2026} \\
\cline{2-4}
&
Truck, vehicle, or mothership coordination &
Cross-platform synchronization explanation &
Joo \textit{et al.}~\cite{joo2026} \\
\cline{2-4}
&
Delivery, monitoring, inspection, and rescue &
Priority-aware dispatch briefing and schedule repair &
Yang \textit{et al.}~\cite{yang2025}, Rigas \textit{et al.}~\cite{rigas2022}, Zhao \textit{et al.}~\cite{zhao2023}, Mao \textit{et al.}~\cite{mao2025}, Pei \textit{et al.}~\cite{pei2023}, Betti Sorbelli \textit{et al.}~\cite{bettisorbelli2025} \\
\hline
\vcenterrow{3}{-1.8ex}{Scalable solvers} &
Deep reinforcement learning assignment &
Policy explanation and compliance-aware critic &
Oubbati \textit{et al.}~\cite{oubbati2021}, Moon \textit{et al.}~\cite{moon2021}, Zhu \textit{et al.}~\cite{zhu2021,zhu2023transferDrlSurveyTPAMI}, Hu \textit{et al.}~\cite{hu2023moMixTPAMI}, Mao \textit{et al.}~\cite{mao2025}, Gong \textit{et al.}~\cite{Gong_2026} \\
\cline{2-4}
&
Metaheuristics and hybrid optimization &
Solver selection, heuristic repair, and black-box search guidance &
Rinaldi \textit{et al.}~\cite{rinaldind}, Zai \textit{et al.}~\cite{zaind}, Mao \textit{et al.}~\cite{mao2025}, Qin and Pournaras~\cite{qin2025strategic} \\
\hline
\vcenterrow{3}{-4ex}{Security and privacy} &
Secure trajectory, relay, and jamming &
RAG-grounded rule checking and secrecy rationale &
Kang \textit{et al.}~\cite{kang2022}, Zhang \textit{et al.}~\cite{zhang2024secureRelay}, Dang-Ngoc \textit{et al.}~\cite{dangngoc2022}, Tedeschi \textit{et al.}~\cite{tedeschi2023} \\
\cline{2-4}
&
Data, video, and access privacy &
Sensitive-context filtering and access-audit explanation &
George \textit{et al.}~\cite{george2023}, Tedeschi \textit{et al.}~\cite{tedeschi2023}, Huang \textit{et al.}~\cite{huang2024} \\
\cline{2-4}
&
Federated, blockchain, and location privacy &
Privacy-aware prompt construction and traceable logs &
Akram \textit{et al.}~\cite{akram2025}, Pu~\cite{pu2023}, Huang \textit{et al.}~\cite{huang2024}, Jagarlapudi \textit{et al.}~\cite{jagarlapudi2022}, Enayati \textit{et al.}~\cite{enayati2023} \\
\hline
\end{tabular}
\end{table*}

The optimization core still enforces feasibility, but the surrounding LLM layer improves the ability to handle open-ended requests, incomplete operator input, and cross-layer tradeoffs \cite{lewis2020rag,shen2023hugginggpt,li2025llmIsacUav,hu2023moMixTPAMI,romera2024funsearch}.

\subsection{Mobility--Communication Assignment}
\label{subsec:assignment_mobility_comm}

The first class in Table~\ref{tab:assignment_taxonomy} appears in UAV-assisted wireless networks, where the central dispatcher must decide not only where UAVs should fly but also which users, relays, sensors, or edge devices they should serve.
In this setting, $\mathbf{P}$, $\mathbf{R}$, and $\mathbf{A}$ are tightly coupled: changing a UAV trajectory changes channel quality, which in turn changes feasible communication schedules, power allocation, and user association.
The assignment problem therefore evolves from geometric
path planning into joint mobility--resource control.

Recent communication-oriented studies establish this coupling through trajectory, access, and resource scheduling.
Liu and Zheng~\cite{liu2022} optimize time-constrained data collection in UAV-enabled environmental monitoring, He \textit{et. al.}~\cite{he2023} connect multi-UAV three-dimensional trajectories with offloading fairness, and Hu
\textit{et. al.}~\cite{hu2024} jointly consider grouping, trajectory, and resource allocation in aerial--ground collaborative computing.
ISAC-oriented work further couples mobility with sensing and communication quality through user association, beamforming, and target localization \cite{jing2024,zhang2024isac,wu2023}.
Across these works, the central dispatcher serves as the integration point where wireless links, sensing objectives, and UAV motion are made consistent.

Data collection and mobile edge computing make the coupling broader because service feasibility depends on both contact time and computing or sensing load.
Nguyen \textit{et. al.}~\cite{nguyen2022} use deep reinforcement learning for three-dimensional trajectory and data collection, Zhu \textit{et. al.}~\cite{zhu2023} introduce a Transformer-based trajectory planner for age-of-information-aware IoT collection, and Shi \textit{et. al.}~\cite{shi2021} embed hierarchical reinforcement learning into trajectory planning and resource allocation.
The common thread is that tasks cannot be assigned independently of communication or computation resources.
Work on transfer learning in deep reinforcement learning further supports the need to reuse learned dispatch policies across changing environments \cite{zhu2023transferDrlSurveyTPAMI}.

Recent work further expands the set of coupled decisions by introducing sensing, ISAC, and LLM-guided multi-objective search.
Zhang \textit{et. al.}~\cite{zhang2024isac} coordinate trajectory, user association, and beamforming for multi-UAV ISAC systems.
Wu \textit{et. al.}~\cite{wu2023} analyze how sensing quality and communication utility interact in trajectory design.
Li \textit{et. al.}~\cite{li2025llmIsacUav} use LLM reasoning to guide multi-objective UAV-ISAC optimization.
Multi-objective multi-agent decision-making provides a complementary reference point for coupled cooperative dispatch \cite{hu2023moMixTPAMI}.
This progression shows why global assignment dispatch is attractive in communication-heavy missions: the central view can expose tradeoffs that are difficult to capture if trajectory, association, and resource allocation are solved separately.

LLM-enabled assignment changes this class from a purely numerical optimization problem into a semantic-to-numerical translation problem.
The central dispatcher may receive a request such as ``prioritize emergency users while keeping sensing accuracy above a minimum level''.
The LLM layer can retrieve policy or service documents, map the request to association, power, sensing, and trajectory constraints, and then select an optimization routine.
Li \textit{et. al.}~\cite{li2025llmIsacUav} provide a concrete UAV-ISAC example in which LLM reasoning is used inside a decomposition-based multi-objective evolutionary algorithm, and related multi-agent decision-making work shows how cooperative objectives can be learned jointly \cite{hu2023moMixTPAMI}.
This suggests that, for mobility--communication assignment, the LLM should act as a trade-off interpreter and search guide, while channel models and optimization solvers remain responsible for feasibility and Pareto quality.

\subsection{Routing, Scheduling, and Vehicle-Assisted Operations}
\label{subsec:assignment_routing_scheduling}

The second class shifts the central variable from wireless link quality to service ordering.
In delivery, inspection, monitoring, and emergency response, the dispatcher mainly determines which UAV serves each task, in what sequence, and under which time, capacity, energy, or vehicle coordination constraints. These problems still use the same decision variables from Section~\ref{sec:preliminaries}, but the dominant coupling is now among $\mathbf{X}$, $\mathbf{P}$, and $\mathbf{S}$.

    Recent routing formulations provide the foundation. Montemanni \textit{et. al.}~\cite{montemannind}, and Joo \textit{et. al.}~\cite{joo2026} study parallel UAV scheduling or drone-delivery routing variants with synchronization and endurance constraints.
Their models extend TSP-style routing with UAV-specific requirements, showing how centralized assignment can convert multiple UAV sorties into a coordinated service sequence rather than a collection of independent paths.

Vehicle-assisted systems add a moving depot, rider, or mothership to the same routing logic.
Robust drone-vehicle routing with uncertain time windows further shows why the central dispatcher must synchronize two mobility layers: the slower ground carrier and the faster aerial task executors \cite{joo2026}.

Application-specific scheduling then adds feasibility constraints that cannot be reduced to distance alone. Persistent and on-demand delivery systems consider rolling horizons, demand uncertainty, docking stations, dispatching points, and urban conflicts \cite{yang2025,torabbeigi2021,pachayappan2021,bettisorbelli2025,pei2023}.
Monitoring and inspection systems add spatial, temporal, edge analytics, and charging constraints \cite{rigas2022,zhao2023,liundInspection,khochare2021,khochare2024}.
Emergency delivery and rescue further prioritize response time and mission urgency \cite{campuzano2022,gongnd}.
These studies form a clear progression: basic routing chooses a feasible service sequence, vehicle-assisted dispatch synchronizes heterogeneous carriers, and application-driven scheduling embeds domain constraints such as monitoring windows, emergency priority, battery exchange, and airspace conflict avoidance.

In LLM-era routing, the difficult step often precedes the VRP or TSP solver.
Natural-language orders, emergency reports, inspection checklists, and operator preferences must be converted into time windows, priorities, vehicle compatibility rules, no-fly constraints, and synchronization requirements.
Tool-oriented LLM agents provide a useful abstraction here: the language model plans which solver or data source to call, executes the tool sequence, and summarizes the result for the operator \cite{shen2023hugginggpt,yao2023react}.
This does not remove the need for exact or heuristic routing algorithms.
Instead, it turns global assignment dispatch into a human-readable loop in which constraints can be inspected before the route is committed.

\subsection{Scalable Solvers and Learning-Enhanced Assignment}
\label{subsec:assignment_scalable_solvers}

Scalable solver design explains how the central assignment can be computed when exact optimization becomes too expensive.
Large UAV dispatch models often combine integer assignment, non-convex trajectory constraints, communication coupling, and time windows.
As the fleet size or task set grows, many studies complement or approximate exact solvers with learning-based or hybrid optimization methods.

Deep reinforcement learning is useful when the dispatcher repeatedly faces dynamic states.
Oubbati \textit{et. al.}~\cite{oubbati2021} adapt UAV base-station positions to urban vehicular traffic, Moon \textit{et. al.}~\cite{moon2021} learn multi-UAV trajectories for target tracking, and Zhu \textit{et. al.}~\cite{zhu2021} jointly learn trajectory planning and cluster-head selection in wireless sensor networks.
Mao \textit{et. al.}~\cite{mao2025} further use a double-level DRL structure for large-scale multi-UAV task scheduling.
These methods keep a centralized coordination objective but use learned policies to avoid solving a full optimization problem from scratch at every decision epoch.

Hybrid solvers are more common when feasibility rules are explicit, but the search space is combinatorial.
Rinaldi \textit{et. al.}~\cite{rinaldind} use a double-chromosome genetic scheduler with safe route planning and Zai \textit{et. al.}~\cite{zaind} combine DRL and particle swarm optimization for emergency material transportation.
Mao \textit{et. al.}~\cite{mao2025} further show how double-level DRL can scale multi-UAV task scheduling, illustrating the practical trend toward solvers that mix prediction, search, and local repair.

LLM-guided search gives this solver layer a new form. Rather than using the LLM to output an assignment directly, the system can let the LLM propose decompositions, priority functions, neighborhood moves, or repair rules, and then score them with an external evaluator.
FunSearch demonstrates this evaluator-guarded pattern for hard algorithmic problems by pairing LLM-generated programs with systematic scoring \cite{romera2024funsearch}.
Toolformer and graph/tree-of-thought reasoning further indicate how LLMs can select external tools or branch over candidate decompositions before committing to a solver \cite{schick2023toolformer,yao2023treeOfThoughts,besta2024graphOfThoughts}.
Self-refinement mechanisms can also revise candidate outputs through feedback \cite{madaan2023selfRefine}.
For UAV assignment, this points to a practical architecture: classical solvers produce or verify feasible dispatch plans, while an LLM searches over solver configurations, local repair strategies, and explanations of why a candidate assignment failed.

The advantage of these methods is speed and adaptability; the cost is weaker guarantees.
Reward design, training distribution, heuristic parameters, and scenario-specific assumptions can strongly affect the quality and safety of the assignment.
Scalable solver design is an implementation dimension of global assignment dispatch rather than a separate dispatch paradigm. It does not change the dispatch variables, but determines whether a centralized model can be solved in real time.

\subsection{Security, Privacy, and Trust-Aware Assignment}
\label{subsec:assignment_security_privacy}

Centralized assignment improves global coordination, but it also concentrates sensitive information. UAV states, user locations, task requests, video streams, trajectories, and communication metadata may all be collected by the dispatcher.
Security- and privacy-aware assignment therefore appears as both an optimization objective and a system-design requirement.

At the physical layer, security is often folded directly into trajectory and resource assignment.
Kang \textit{et. al.}~\cite{kang2022} coordinate dual-UAV trajectories and power for bidirectional secure communication.
Zhang \textit{et. al.}~\cite{zhang2024secureRelay} address collaborative secure relay communications in UAV swarms, while Dang-Ngoc \textit{et. al.}~\cite{dangngoc2022} use cooperative friendly jamming to protect swarm-assisted communications.
These methods treat secrecy not as a post-processing rule, but as a dispatch constraint coupled with $\mathbf{P}$, $\mathbf{R}$, and $\mathbf{A}$.

Data privacy and access control address a different part of the centralized risk.
George \textit{et. al.}~\cite{george2023} develop multi-party access control for UAV services, and Tedeschi \textit{et. al.}~\cite{tedeschi2023} protect collision-avoidance coordination without exposing unnecessary motion information.
The dispatch logic here is organizational as much as mathematical: the central platform must decide who may see which task, data stream, or UAV state.

Distributed privacy mechanisms reduce dependence on raw centralized data.
Akram \textit{et. al.}~\cite{akram2025} use conditional federated learning for post-disaster infrastructure monitoring.
Pu~\cite{pu2023} studies privacy-preserving and
fault-tolerant data aggregation, and Huang \textit{et. al.}~\cite{huang2024} analyze privacy-aware remote
identification and routing.
Location privacy studies such as Jagarlapudi \textit{et. al.}~\cite{jagarlapudi2022} and Enayati \textit{et. al.}~\cite{enayati2023} further show that assignment plans and trajectories themselves may leak sensitive information.
These works show that centralized dispatch should optimize performance together with data minimization, access control, and location obfuscation.

The LLM layer intensifies these risks because prompts and retrieved contexts may contain user identities, mission intent, facility locations, or predicted trajectories.
A privacy-aware meta-dispatcher therefore needs explicit filters before prompt construction, access control over retrieved documents, and audit logs for tool calls and generated explanations.
RAG helps ground decisions in external documents, but it also makes provenance and document freshness part of the security boundary \cite{lewis2020rag}.
Thus, trustworthy global assignment in the LLM era requires joint control of optimization variables and information exposure.

\section{Decentralized Agentic Dispatch}
\label{sec:grab_go_dispatch}

\begin{figure}[t]
    \centering
    \includegraphics[width=0.85\columnwidth]{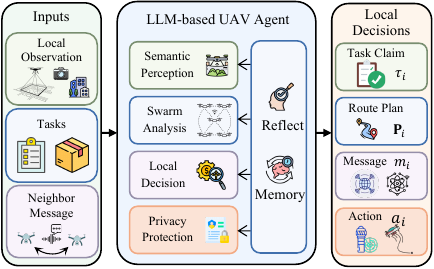}
    \caption{Decentralized agentic dispatch lets local UAV agents select tasks from observations, utilities, short-term memory, and limited peer messages.}
    \label{fig:grab_go_dispatch}
\end{figure}

Decentralized agentic dispatch describes a paradigm in which UAVs select tasks, adjust trajectories, and coordinate with nearby agents from local observations, local utility, compact semantic messages, and limited peer negotiation rather than from a complete centralized assignment.
In the notation of Section~\ref{sec:preliminaries}, the same decisions $\mathbf{X}$, $\mathbf{P}$, $\mathbf{S}$, $\mathbf{R}$, and $\mathbf{A}$ may still be involved, but they are produced through local policies, bidding, consensus, or distributed learning. Table~\ref{tab:grabgo_taxonomy} organizes this paradigm from the lowest level of autonomous motion to swarm interaction, task ownership, and privacy-preserving autonomy.
In the LLM era, the local agent can summarize nearby events, maintain short-term task memory, generate negotiation messages, reflect on failed local actions, and pass only verified high-level commands to conventional planners or controllers \cite{driess2023palmE,lin2025navcotTPAMI,yao2023react,shinn2023reflexion}.
Figure~\ref{fig:grab_go_dispatch} illustrates this decentralized agentic logic.

\begin{table*}[t]
\centering
\caption{Taxonomy of representative decentralized agentic UAV dispatch methods and their LLM-agent roles.}
\label{tab:grabgo_taxonomy}
\footnotesize
\setlength{\arrayrulewidth}{0.4pt}
\setlength{\tabcolsep}{2pt}
\renewcommand{\arraystretch}{1.08}
\begin{tabular}{|>{\raggedright\arraybackslash}m{0.22\textwidth}
                |>{\raggedright\arraybackslash}m{0.22\textwidth}
                |>{\raggedright\arraybackslash}m{0.24\textwidth}
                |>{\raggedright\arraybackslash}m{0.28\textwidth}|}
\hline
\textbf{Agentic Layer} &
\textbf{Local Dispatch Role} &
\textbf{LLM-Agent Role} &
\textbf{Representative Methods} \\
\hline
\vcenterrow{3}{-3ex}{Local decision layer} &
Reinforcement-learning trajectory control &
Perception--memory--action loop for local intent interpretation &
Ebrahimi \textit{et al.}~\cite{ebrahimi2021}, Nguyen \textit{et al.}~\cite{nguyen2022}, Zhu \textit{et al.}~\cite{zhu2023,zhu2023transferDrlSurveyTPAMI} \\
\cline{2-4}
&
Active perception under uncertainty &
Multimodal observation summarizer and next-view planner &
Zhu and Chen~\cite{zhu2024openVocabularySurveyTPAMI}, Zhu \textit{et al.}~\cite{zhu2025llafsTPAMI}, Kirillov \textit{et al.}~\cite{kirillov2023segmentAnything} \\
\cline{2-4}
&
Tracking, avoidance, and covert motion &
Local safety critic and intent-aware motion proposal &
Xu \textit{et al.}~\cite{xu2022}, Zhang \textit{et al.}~\cite{zhang2023}, Huang \textit{et al.}~\cite{huang2022} \\
\hline
\vcenterrow{3}{-3ex}{Swarm coordination layer} &
Decentralized coverage and exploration &
Peer-agent message compression and local consensus prompting &
{blue}Zhao \textit{et al.}~\cite{zhao2021}, Huang \textit{et al.}~\cite{huang2025}\\
\cline{2-4}
&
Relative-state UAV interaction &
Relative-observation grounding for semantic communication &
Hasan \textit{et al.}~\cite{hasan2023} \\
\cline{2-4}
&
MARL and distributed reinforcement learning &
Role-aware multi-agent reasoning with bounded messages &
Venturini \textit{et al.}~\cite{venturini2020distributed}, Qin and Pournaras~\cite{qin2025strategic}, Hu \textit{et al.}~\cite{hu2023moMixTPAMI}, Alkahtani \textit{et al.}~\cite{alkahtani2026explainable}, Luo \textit{et al.}~\cite{luo2025multiLlmEdge} \\
\hline
\vcenterrow{3}{0ex}{Task ownership layer} &
Consensus-based auctions &
Bid explanation and task-claim negotiation &
Mao \textit{et al.}~\cite{mao2025} \\
\cline{2-4}
&
Dynamic partial replanning &
Reflection-based local repair after state changes &
Qin and Pournaras~\cite{qin2025strategic} \\
\hline
\vcenterrow{3}{-2ex}{Privacy and trust layer} &
Privacy-aware or covert trajectories &
Sensitive-context filtering before local reasoning &
Huang \textit{et al.}~\cite{huang2022}, Chinthi-Reddy \textit{et al.}~\cite{chinthi-reddy2022} \\
\cline{2-4}
&
Private trajectory exchange and collision avoidance &
Minimal semantic message sharing among agents &
Sciancalepore and George~\cite{sciancalepore2022}, Tedeschi \textit{et al.}~\cite{tedeschi2023} \\
\cline{2-4}
&
Private computation and federated learning &
Local memory isolation and federated agent learning &
Gao \textit{et al.}~\cite{gaond}, Akram \textit{et al.}~\cite{akram2025} \\
\hline
\end{tabular}
\end{table*}

\subsection{Local Perception and Motion Decisions}
\label{subsec:grabgo_local_motion}

Decentralized agentic dispatch begins with the individual UAV's ability to turn local perception into motion and task-selection decisions.
Unlike global assignment dispatch, where a central node decides the next task or route, the UAV locally evaluates whether a target is reachable, valuable, safe, or informative.
This makes local trajectory control the first layer of the taxonomy in Table~\ref{tab:grabgo_taxonomy}.

Agent-level reinforcement learning provides a natural mechanism for this local decision layer.
Ebrahimi \textit{et. al.}~\cite{ebrahimi2021} learn autonomous UAV trajectories for ground-object localization while considering energy, trajectory length, waypoints, and flight time.
Nguyen \textit{et. al.}~\cite{nguyen2022} use deep reinforcement learning for three-dimensional trajectory and data-collection optimization, allowing the UAV to adapt its route and collection time online.
Zhu \textit{et. al.}~\cite{zhu2023} introduce a transformer-based trajectory planner for age-of-information-aware data collection, showing that sequence models can help local agents account for temporal freshness when deciding where to fly next.
These studies share the same dispatch logic: the UAV does not wait for a complete route, but repeatedly maps current observations to an executable motion decision.
In addition, a survey on transfer learning in deep reinforcement learning further highlights why reusable policies matter when local UAV agents move across missions, weather, or sensing domains \cite{zhu2023transferDrlSurveyTPAMI}.

Beyond learned local policies, recent semantic aerial navigation systems suggest that memory and behavior-tree structures can improve resilience by linking perception, memory, and action selection \cite{kim2025raven}.
In an LLM-enabled agent, a multimodal LLM can further summarize what the UAV sees, infer whether a local target matches the mission intent, and pass a structured goal to a verified planner.
PaLM-E demonstrates how language and embodied observations can be represented in a single
reasoning interface \cite{driess2023palmE}.
UAV-specific benchmarks such as AerialVLN \cite{liu2023aerialvln} and AeroDuo \cite{wu2025aeroduo} show that aerial agents must ground instructions in altitude, viewpoint, and long-range spatial relations, while broader vision-language navigation studies \cite{lin2025navcotTPAMI,chen2025constraintAwareVLNTPAMI,wang2026imagineNavTPAMI,lin2026evolveNavTPAMI} provide mechanisms for constraint-aware and reasoning-based navigation.
With such semantic mediation, the local decision layer goes beyond observation-to-action mapping and instead transforms observations into semantic states and verified goals.
This semantic mediation also enables the agent to explicitly reason about uncertainty, which naturally extends local decision-making from action execution to information-seeking motion.

Extending this semantic and uncertainty-aware decision framework, UAVs further engage in active perception, where movement is selected not only for task completion but also for information gain.
Xu \textit{et. al.}~\cite{xu2022} combine obstacle avoidance and target tracking through DRL, Zhang \textit{et. al.}~\cite{zhang2023} study online pursuit-evasion planning for multiple UAVs, and Huang \textit{et. al.}~\cite{huang2022} plan covert surveillance trajectories for mobile targets.
Open-vocabulary, few-shot, and foundation segmentation models also matter here because they make local targets and region extraction more adaptable before the motion policy is invoked \cite{zhu2024openVocabularySurveyTPAMI,zhu2025llafsTPAMI,kirillov2023segmentAnything}.
This progression suggests that, as uncertainty and adversarial movement increase, local autonomy must jointly reason about sensing value, safety, and mission utility.

\subsection{Swarm Interaction and Distributed Learning}
\label{subsec:grabgo_swarm_learning}

Once multiple UAVs make local decisions, the main problem becomes coordination.
Purely independent agents may duplicate coverage, collide, or chase the same target.
The second layer of Table~\ref{tab:grabgo_taxonomy} therefore focuses on local interaction: UAVs still act autonomously, but neighboring agents exchange limited information to make the swarm behavior coherent.
Decentralized quadrotor swarm systems such as EGO-Swarm further demonstrate that local planning, inter-agent collision avoidance, and limited information exchange can support fully autonomous swarm operation in cluttered environments \cite{zhou2021ego}.

Coverage and exploration methods show the basic form of this interaction.
Experimental platforms such as AirSwarm also highlight the importance of low-cost, reproducible multi-UAV systems for validating decentralized coordination, communication, and task-allocation algorithms in physical environments \cite{li2025airswarm}.
Zhao \textit{et. al.}~\cite{zhao2021} use decentralized learning for energy-efficient multi-UAV content coverage. Huang \textit{et. al.}~\cite{huang2025} design biomimetic swarm exploration with UAV-to-UAV communication.
These methods move beyond single-agent motion by letting each UAV adapt to local coverage gaps, neighbor behavior, and changing task ownership.
Resource-constrained micro-UAV swarms further show that decentralized exploration can be achieved with minimal sensing and communication, which is important for dispatch policies under onboard compute, energy, and payload limits \cite{mcguire2019minimal}.

Relative-state interaction then makes the swarm less dependent on global coordinates or centralized maps.
Hasan \textit{et. al.}~\cite{hasan2023} study UAV-to-UAV interaction through three-dimensional relative-direction-based evolutionary computation.
This type of local relation is important in low-altitude operations, where GPS errors, communication limits, and obstacle-rich environments may make a fully shared global state unreliable.

Multi-agent reinforcement learning generalizes these ideas into a learned coordination policy.
Venturini \textit{et. al.}~\cite{venturini2020distributed} use distributed reinforcement learning for flexible swarm control, and Qin and Pournaras~\cite{qin2025strategic} combine short-term distributed optimization with long-term reinforcement learning.
Multi-objective multi-agent cooperative decision-making further connects local learning with coupled team objectives \cite{hu2023moMixTPAMI}, while Alkahtani
\textit{et. al.}~\cite{alkahtani2026explainable} add explainability and trust evaluation to secure FANET routing.
Together these works show that decentralized agentic dispatch can scale from local motion to swarm-level coordination when agents learn not only what action is good for themselves, but also how that action affects nearby UAVs.

For LLM-enabled swarms, the communication payload changes from raw state vectors to compact semantic messages.
A UAV may report that a road segment is blocked, a target is uncertain, or a relay role is needed, and neighboring agents can reason over this message without sharing full video streams or full trajectories.
Multi-LLM edge orchestration provides a useful model for this setting because different agents can specialize in perception, planning, communication, or trust assessment while remaining constrained by edge latency and privacy budgets \cite{luo2025multiLlmEdge}.
The challenge is to make semantic communication bounded and verifiable: messages should be short enough for local links, grounded enough to avoid hallucination, and structured enough for consensus or safety modules to consume.

\subsection{Bidding, Consensus, and Partial Replanning}
\label{subsec:grabgo_bidding_consensus}

Learning-based coordination is flexible, but task ownership often needs a more explicit mechanism.
Bidding and consensus methods provide this mechanism by letting UAVs evaluate task utilities, communicate bids, and agree on winners.
A representative example is variable-swarm distributed auction assignment combined with sequential convex programming, where task ownership and trajectory feasibility are jointly handled through auction-based allocation and trajectory optimization \cite{doi:10.2514/6.2015-0599}.
This mechanism directly captures decentralized task claiming: a UAV obtains ownership because its local utility is competitive, rather than because a central dispatcher assigns the task.

Recent work adapts consensus and bidding structures to more dynamic and heterogeneous UAV systems.
Mao \textit{et. al.}~\cite{mao2025} use a double-level DRL structure for large-scale multi-UAV task scheduling, so learned policies can support local reassignment without rebuilding the entire allocation.
For UAV dispatch, the importance is not only the auction format but the structure it imposes: local cost estimates such as travel distance, energy, deadline, risk, and task reward become the bridge between individual decisions and swarm-level task ownership.

LLM agents can strengthen this ownership layer by making bids and replanning decisions explainable.
A local agent can translate its energy state, distance, risk, and sensor capability into a bid rationale.
It could ask neighboring agents for missing context and reflect on why a previous claim failed \cite{yao2023react,shinn2023reflexion}.
However, the final ownership decision should still be enforced by the auction or consensus protocol.
This separation is necessary because language can improve negotiation and diagnosis, whereas agreement, conflict resolution, and deadline feasibility require explicit distributed rules.

\subsection{Privacy-Preserving and Trustworthy Autonomy}
\label{subsec:grabgo_privacy_trust}

Decentralization reduces dependence on a central server, but it does not remove privacy and security risks.
Instead, the risk moves to peer-to-peer messages, trajectory sharing, collision-avoidance protocols, cooperative learning, and local sensing.
The final layer of Table~\ref{tab:grabgo_taxonomy} therefore treats privacy and trustworthiness as part of autonomous dispatch, not as an external communication add-on.

Some studies embed privacy directly into trajectory behavior.
Huang \textit{et. al.}~\cite{huang2022} plan covert surveillance trajectories that maintain monitoring performance while reducing detectability, and Chinthi-Reddy \textit{et. al.}~\cite{chinthi-reddy2022} propose DarkSky to reduce privacy invasion in target tracking.
These works show that privacy can affect $\mathbf{P}$ itself: the path should serve the task
while avoiding unnecessary exposure of either the UAV or non-target areas.

Other works protect coordination messages.
Sciancalepore and George~\cite{sciancalepore2022} develop privacy-preserving trajectory matching for autonomous UAVs, and Tedeschi \textit{et. al.}~\cite{tedeschi2023} propose PPCA for privacy-preserving collision avoidance.
Their common concern is that safe coordination often requires sharing future motion, yet future motion can reveal sensitive locations or mission intent.
Privacy-aware decentralized agentic dispatch must therefore decide what information is necessary for coordination and what should remain hidden.

Privacy-preserving computation and learning complete the picture.
Gao \textit{et. al.}~\cite{gaond} study UAV-centric privacy-preserving computation offloading in multi-UAV MEC with local differential privacy, while Akram \textit{et. al.}~\cite{akram2025} use conditional federated learning for post-disaster drone services.
These methods point to a privacy-preserving form of autonomy in which UAVs can improve local policies or offloading decisions without revealing raw observations, trajectories, or mission data.

LLM-enabled autonomy adds prompt-level and memory-level privacy risks.
Local summaries may expose sensitive locations even when raw images are not transmitted, and reflective memories may preserve failed routes, targets, or operator instructions longer than necessary.
Therefore, a trustworthy decentralized agent should separate private sensor memory from shareable semantic memory, filter messages before negotiation, and bind every LLM-proposed action to a safety or privacy check.
Privacy-preserving dispatch should therefore cover not only encrypted communication, but also controlled language generation, memory isolation, and minimal disclosure during peer coordination.



\section{Divide-and-Conquer Dispatch}
\label{sec:divide_conquer_dispatch}

\begin{figure}[t]
    \centering
    \includegraphics[width=0.85\columnwidth]{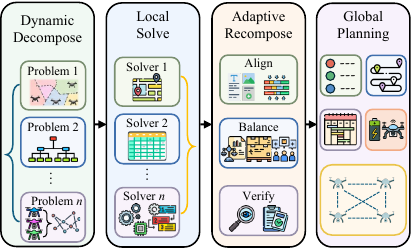}
    \caption{Divide-and-conquer dispatch partitions a large mission into local subproblems, solves them separately, and reconciles them through a coordinator.}
    \label{fig:divide_conquer_dispatch}
\end{figure}

Divide-and-conquer dispatch reduces a large coupled UAV mission by splitting the problem into smaller subproblems across different dimensions such as decisions, objectives, space, time, groups, roles, and communication links.
The resulting subproblems are solved separately and then coordinated through recomposition or verification mechanisms.
Unlike pipeline dispatch, divide-and-conquer dispatch does not assume a fixed module sequence; unlike global assignment dispatch, it does not rely on one global solver.
Its core question is where to draw the boundaries so that trajectory, schedule, resource, association, and security decisions remain locally tractable while still producing a coherent global dispatch plan.
Table~\ref{tab:divide_conquer_taxonomy} summarizes representative methods according to the main decomposition principle they use.
In the LLM era, the decomposition boundary can also be semantic and computational: knowledge may be split across retrieval sources, agent roles may be assigned by capability, memory may be scoped to regions or subteams, and reasoning may be shared between onboard models, edge servers, and cloud models.
This makes divide-and-conquer dispatch a natural bridge between classical decomposition and multi-agent LLM orchestration \cite{shen2023hugginggpt,luo2025multiLlmEdge,yao2023react,besta2024graphOfThoughts}.
Figure~\ref{fig:divide_conquer_dispatch} illustrates this partition-and-coordinate logic.

\begin{table*}[t]
\centering
\caption{Taxonomy of representative divide-and-conquer dispatch methods and their LLM-era orchestration roles.}
\label{tab:divide_conquer_taxonomy}
\footnotesize
\setlength{\arrayrulewidth}{0.4pt}
\setlength{\tabcolsep}{2pt}
\renewcommand{\arraystretch}{1.08}
\begin{tabular}{|>{\raggedright\arraybackslash}m{0.22\textwidth}
                |>{\raggedright\arraybackslash}m{0.22\textwidth}
                |>{\raggedright\arraybackslash}m{0.24\textwidth}
                |>{\raggedright\arraybackslash}m{0.28\textwidth}|}
\hline
\textbf{Decomposition Principle} &
\textbf{Dispatch Role} &
\textbf{LLM-Era Orchestration Role} &
\textbf{Representative Methods} \\
\hline
\vcenterrow{3}{-2.5ex}{Decision and objective decomposition} &
Trajectory, energy, and link tradeoff &
Objective decomposer and Pareto-search guide &
Javed \textit{et al.}~\cite{javed2023}, Meng \textit{et al.}~\cite{meng2022}, Jing \textit{et al.}~\cite{jing2024} \\
\cline{2-4}
&
Secure trajectory, power, and robustness &
Security-rule retrieval and feasibility critic &
Kang \textit{et al.}~\cite{kang2022}, Zhang \textit{et al.}~\cite{zhang2024secureRelay}, Dang-Ngoc \textit{et al.}~\cite{dangngoc2022} \\
\cline{2-4}
&
Temporal, connectivity, and ISAC objectives &
Communication--sensing tradeoff explanation &
Meng \textit{et al.}~\cite{meng2022}, Jing \textit{et al.}~\cite{jing2024}, Zhang \textit{et al.}~\cite{zhang2024isac}, Li \textit{et al.}~\cite{li2025llmIsacUav} \\
\hline
\vcenterrow{3}{-3ex}{Spatial and regional decomposition} &
Three-dimensional and multi-region coverage &
Semantic map and region partitioning &
Xiong \textit{et al.}~\cite{xiong2025}, Li \textit{et al.}~\cite{li2025hierarchical} \\
\cline{2-4}
&
Emergency waypoints and rescue paths &
Disaster-scene grounding and priority partitioning &
Liu \textit{et al.}~\cite{liu202501}, Liu \textit{et al.}~\cite{liu2025llmInspection}, Lin \textit{et al.}~\cite{lin2024correctableLandmarkTPAMI,lin2025navcotTPAMI}, Xiong \textit{et al.}~\cite{xiong2025} \\
\cline{2-4}
&
Inspection and agriculture scene modeling &
Multimodal region labeling and task extraction &
Liu \textit{et al.}~\cite{liu202501}, Liu \textit{et al.}~\cite{liu2025llmInspection}, Zhang \textit{et al.}~\cite{zhang2024vlmSurveyTPAMI}, Zhu and Chen~\cite{zhu2024openVocabularySurveyTPAMI}, Kirillov \textit{et al.}~\cite{kirillov2023segmentAnything} \\
\hline
\vcenterrow{3}{-3ex}{Swarm, role, and function decomposition} &
Grouping and resource collaboration &
Role-specialized multi-agent planner &
Hu \textit{et al.}~\cite{hu2024}, Luo \textit{et al.}~\cite{luo2025multiLlmEdge}, Alkahtani \textit{et al.}~\cite{alkahtani2026explainable} \\
\cline{2-4}
&
Search stages and spatio-temporal planning &
Stage-specific agent memory and progress monitor &
Wu \textit{et al.}~\cite{wu2021}, Gong \textit{et al.}~\cite{Gong_2026} \\
\cline{2-4}
&
Formation, recharging, and farm operations &
Skill-library reuse for repeated sub-missions &
Alyassi \textit{et al.}~\cite{alyassi2023}, Ming \textit{et al.}~\cite{mingnd} \\
\hline
\vcenterrow{3}{-3ex}{Security and communication decomposition} &
Secure relay and friendly jamming &
Policy-aware security sub-agent and audit log &
Zhang \textit{et al.}~\cite{zhang2024secureRelay}, Dang-Ngoc \textit{et al.}~\cite{dangngoc2022} \\
\cline{2-4}
&
Mobility--power robust security &
Adversarial-context summarization and robust repair &
Kang \textit{et al.}~\cite{kang2022}, Zhang \textit{et al.}~\cite{zhang2024secureRelay} \\
\cline{2-4}
&
QoS and relay-aware mission execution &
Communication-state retrieval for local recomposition &
Yanmaz \textit{et al.}~\cite{yanmaz2024} \\
\hline
\end{tabular}
\end{table*}

\subsection{Trajectory--Communication and Objective Decomposition}
\label{subsec:dc_objective_decomposition}

The first class in Table~\ref{tab:divide_conquer_taxonomy} decomposes strongly coupled trajectory-communication objectives.
In UAV networks, mobility affects channel quality, energy consumption, sensing accuracy, secrecy, and delay.
Solving all variables jointly is often non-convex, so many methods separate the problem into trajectory, power, beamforming, time-frame, or objective subproblems and coordinate them iteratively.

Energy- and communication-aware studies illustrate the basic decomposition.
Javed \textit{et. al.}~\cite{javed2023} optimize outage probability and trajectory length as related but distinguishable objectives in disaster scenarios, while recent ISAC work decomposes trajectory, beamforming, association, and target-localization quality \cite{meng2022,jing2024,zhang2024isac}.
These works use decomposition to expose why a path that is short geometrically may not be efficient once communication or sensing performance is considered.

Secure communication makes the objective split sharper.
Kang \textit{et. al.}~\cite{kang2022} coordinate dual-UAV trajectories and transmit power for bidirectional secure communication, Zhang \textit{et. al.}~\cite{zhang2024secureRelay} study collaborative secure relay communication under colluding eavesdroppers, and Dang-Ngoc \textit{et. al.}~\cite{dangngoc2022} decompose relay and friendly jamming functions in swarm UAV-assisted networks.
In these settings, security is not an independent layer after planning.
It changes how mobility, role, and resource variables should be separated and recombined.

Temporal and sensing-oriented decomposition further extends the same idea.
Meng \textit{et. al.}~\cite{meng2022} use frame-based iterative optimization for UAV trajectory and beamforming in integrated periodic sensing and communication.
Jing \textit{et. al.}~\cite{jing2024} design multi-stage UAV trajectories for joint communication and target localization, while Li \textit{et. al.}~\cite{li2025llmIsacUav} show how LLM reasoning can guide the decomposition of multi-objective  UAV-ISAC search.
The progression from energy and rate to secrecy, connectivity, and localization shows that objective decomposition is useful whenever one trajectory must serve several competing system functions.

LLM-era objective decomposition makes the split more explicit to human operators and downstream solvers.
The LLM can translate mission intent into named subobjectives, retrieve relevant constraints, and decide whether a mobility, communication, sensing, privacy, or energy solver should be called first.
In UAV-ISAC optimization, LLM reasoning has already been used to guide multi-objective search \cite{li2025llmIsacUav}.
More generally, evaluator-guarded LLM search, as in FunSearch \cite{romera2024funsearch}, suggests a way to generate candidate decomposition rules or local repair programs while retaining
external scoring.
Thus, the LLM contribution is not to merge all objectives into one opaque prompt, but to expose and coordinate the subproblems that must still be verified numerically.

\subsection{Spatial and Regional Decomposition}
\label{subsec:dc_spatial_regional}

The second class partitions the operating environment or task area.
This is especially natural in coverage, inspection, agriculture, construction, and disaster response, where the mission spans a large region and each subregion can be planned locally.
Spatial decomposition changes the dispatch question from ``What is the best global route?'' to ``How should the space be divided, and how should local routes be coordinated across boundaries?''
Frontier- or coverage-guided planners further support this view by decomposing unknown environments into informative regions and then coordinating local exploration trajectories across spatial boundaries \cite{feng2022fcplanner}.

Coverage planning provides the clearest example. Xiong \textit{et. al.}~\cite{xiong2025} optimize collaborative coverage paths for multi-region post-disaster assessment, using region partitioning, task allocation, and coverage routing to improve swarm efficiency.
Li \textit{et. al.}~\cite{li2025hierarchical} similarly decouple multi-region coverage while incorporating task-priority awareness.
These methods show that spatial decomposition is not merely a computational simplification.
It also defines which UAV is responsible for which part of the mission.

Emergency and domain-specific missions add priorities and sensing requirements to the spatial split.
Large-scale exploration in GPS-denied cave environments further demonstrates that spatial decomposition must be coupled with mapping, localization, communication constraints, and local replanning when the global state is incomplete \cite{petravcek2021large}.
Liu \textit{et. al.}~\cite{liu202501} use 3D reconstruction to guide refinement inspection at construction sites, Liu \textit{et. al.}~\cite{liu2025llmInspection} convert inspection evidence into multi-agent subtasks, and Xiong \textit{et. al.}~\cite{xiong2025} partition post-disaster assessment regions for collaborative coverage.
In these studies, the environment model becomes part of the decomposition: better region or scene understanding leads to better local dispatch subproblems.

Spatial decomposition is also where multimodal LLMs become directly relevant to UAV dispatch.
A region may be described by satellite imagery, onboard video, maps, inspection manuals, and human incident reports.
An LLM or multimodal LLM can label zones, extract candidate targets, and produce region-level task descriptions before a coverage or routing solver runs. PaLM-E provides a general embodied multimodal
precedent \cite{driess2023palmE}.
Recent VLM studies and benchmarks clarify how visual-language grounding should be evaluated \cite{zhang2024vlmSurveyTPAMI,xu2025lvlmEhubTPAMI,yang2026defyingDistractionsTPAMI,zhang2026beyondLlavaHdTPAMI}.
Open-vocabulary and few-shot segmentation studies support region-level perception \cite{zhu2024openVocabularySurveyTPAMI,zhu2025llafsTPAMI}.
AerialVLN, AeroDuo, and large-model landmark discovery adapt language grounding to navigation \cite{liu2023aerialvln,wu2025aeroduo,lin2024correctableLandmarkTPAMI,lin2025navcotTPAMI,chen2025constraintAwareVLNTPAMI}.
LLM-informed drone inspection further shows how textual and visual evidence can be turned into multi-agent inspection subtasks
\cite{liu2025llmInspection}.
Therefore, spatial decomposition in LLM-enabled dispatch is not merely geometric partitioning; it also involves semantic region construction followed by verifiable local planning.

Spatial decomposition is effective but not free.
Boundaries between regions may create duplicated coverage, missed targets, relay gaps, or conflicts over shared charging stations and air corridors.
Therefore, the success of this class depends on both the partition rule and the coordination rule that reconnects local routes into a global mission.

\subsection{Swarm, Role, and Functional Decomposition}
\label{subsec:dc_swarm_role}

The third class decomposes the UAV fleet itself.
Instead of assigning all UAVs to a single large optimization problem, the system forms groups, roles, or functional modules.
This approach is useful when UAVs have heterogeneous payloads, communication ranges, endurance levels, or mission responsibilities.

Group-level decomposition appears in aerial--ground computing and networked service systems.
Hu \textit{et. al.}~\cite{hu2024} jointly optimize UAV grouping, trajectories, and resource allocation for swarm-assisted aerial--ground collaborative computing.
Luo \textit{et. al.}~\cite{luo2025multiLlmEdge} extend this idea to multi-LLM edge orchestration, where model roles and placement must be matched to communication and trust constraints.
These works use grouping to reduce coordination dimension while preserving enough structure for resource allocation.

Search and urban-airspace planning show another form of functional decomposition.
Wu \textit{et. al.}~\cite{wu2021} study swarm-based four-dimensional path planning, decomposing dense urban coordination over space and time to reduce conflicts.
Gong \textit{et. al.}~\cite{Gong_2026} add an LLM-aware compliance layer for low-altitude trajectory planning, making regulatory constraints part of the
decomposition.
Here, the subproblems are both group-specific and stage-specific, separating mission stages from time-dependent safety constraints.

Role-based decomposition is especially useful in long-duration or hostile missions. 
Alyassi \textit{et. al.}~\cite{alyassi2023} decompose persistent UAV missions into flight execution and autonomous recharging decisions.
Ming \textit{et. al.}~\cite{mingnd} compare swarm control mechanisms for unmanned farms, highlighting the separation of navigation, perception, visual collaboration, and trajectory planning.
These examples show that divide-and-conquer dispatch can be organizational: the mission is simplified by deciding which UAV or subsystem owns which function.

LLM-agent systems make this organizational view more concrete.
A fleet can be decomposed into role-specific agents: a perception agent summarizes aerial observations, a planning agent proposes routes, a communication agent checks connectivity, a privacy agent filters messages, and a verifier rejects unsafe proposals.
HuggingGPT provides a general tool-orchestration pattern for assigning subtasks to specialized models \cite{shen2023hugginggpt}, while multi-LLM edge intelligence studies how such roles can be placed across devices, edge servers, and cloud resources under trust and latency constraints \cite{luo2025multiLlmEdge}.
For UAV dispatch, role decomposition therefore includes both physical roles among UAVs and computational roles among models.

\subsection{Security- and Communication-Oriented Decomposition}
\label{subsec:dc_security_comm}

Security and communication can also define the decomposition boundary.
In many missions, some UAVs should serve as relays, some should jam eavesdroppers, some should sense or collect data, and others should maintain connectivity.
Decomposing these roles allows the dispatch system to optimize specialized subproblems while maintaining a shared security or communication objective.

Secure relay and friendly jamming provide direct examples.
Zhang \textit{et. al.}~\cite{zhang2024secureRelay} study UAV-swarm-enabled collaborative secure relay communication under colluding eavesdroppers, coordinating relay trajectories and resources to improve secrecy capacity.
Dang-Ngoc \textit{et. al.}~\cite{dangngoc2022} separate relay communication and cooperative friendly jamming in swarm UAV-assisted networks.
These methods are divide-and-conquer in a functional sense: communication support and security protection are assigned to different roles and then coordinated.

The mobility-power decomposition discussed earlier also becomes more important under adversarial or uncertain channels.
Kang \textit{et. al.}~\cite{kang2022} and Zhang \textit{et. al.} ~\cite{zhang2024secureRelay} show that trajectory, power control, relay role, and robustness can be optimized as connected subproblems rather than one opaque secure-dispatch model.
This separation clarifies which part of the solution improves channel geometry, which part controls signal exposure, and which part protects against uncertainty.

Communication-oriented decomposition is also visible in rescue and hostile-environment planning.
Yanmaz \textit{et. al.}~\cite{yanmaz2024} separate search and relay roles to maintain connectivity during multi-target search.
It suggests that communication should not be treated only as a constraint checked after planning; it can be a decomposed dispatch role that shapes where UAVs fly and how the mission is divided.

LLM orchestration adds another security boundary.
A security sub-agent may retrieve airspace or communication policies, summarize adversarial context, or explain why a relay or jamming role is needed, but it can also introduce prompt injection, stale evidence, or overconfident recommendations.
ReAct-style tool use is helpful only if every retrieved rule, generated explanation, and proposed role assignment is logged and checked by a non-LLM verifier \cite{yao2023react}.
Thus, security-oriented decomposition must cover both radio and trajectory security and the trustworthiness of the language-mediated coordination process.



\section{Synthesis and Future Directions in the LLM Era}
\label{sec:opportunities_challenges}

Sections~\ref{sec:pipeline_dispatch}--\ref{sec:divide_conquer_dispatch}
reviewed four dispatch paradigms and embedded LLM-related mechanisms
inside each paradigm. Read together, these paradigms suggest that LLMs
are not simply an additional module placed after optimization or
control. They act across the dispatch stack, linking operator intent,
mission knowledge, multimodal perception, tool invocation, solver
selection, and human oversight. Optimization, control, communication
theory, and safety verification still define the physical limits of a
mission; the LLM layer changes how these components are selected,
connected, explained, and adapted during real operations
\cite{lewis2020rag,shen2023hugginggpt,driess2023palmE,luo2025multiLlmEdge}.

\subsection{Cross-Paradigm Synthesis}
\label{subsec:llm_synthesis}

The four paradigms remain useful abstractions, but their boundaries
become less rigid once semantic reasoning is introduced. A practical
LLM-enabled dispatch system may begin as a pipeline when an operator
describes a mission, behave like a global assignment system when fleet
resources must be matched to tasks, shift toward agentic
Grab-and-Go when nearby UAVs react to local events, and rely on
divide-and-conquer when the mission becomes too large for a single
planner. The dispatch paradigm is therefore not only a static design
choice; it can also become a mode selected according to mission scale,
connectivity, uncertainty, and safety requirements.

Because pipeline dispatch already exposes intermediate artifacts, it
can absorb LLMs without hiding the intermediate decisions.
Natural-language and multimodal inputs can be translated into task
graphs, retrieval calls, solver inputs, trajectory candidates, and
monitoring reports, with each artifact checked before the next stage
executes
\cite{lewis2020rag,shen2023hugginggpt,Gong_2026}. Global assignment uses the same semantic layer at a different level of abstraction.
When a central dispatcher faces heterogeneous objectives, an LLM can summarize fleet state, retrieve relevant rules, compare competing assignments, and make the reasoning behind a route or association legible to a human operator.
The actual assignment still has to be produced or verified by explicit optimization and constraint models, since global feasibility cannot be inferred from linguistic plausibility alone \cite{li2025llmIsacUav}.

When dispatch becomes decentralized, the LLM is no longer only an interpreter for a central plan. In decentralized agentic dispatch, an
edge-hosted or lightweight onboard LLM can interpret nearby events,
compress negotiation messages, and help a UAV decide whether to claim,
defer, or reject a task. This is valuable under partial observation,
yet it also makes latency, message size, local memory, and
prompt-level attack surfaces part of the dispatch design rather than
secondary implementation details
\cite{liu2025llmInspection,xu2025lvlmEhubTPAMI,wang2026vlBiasBenchTPAMI,luo2025multiLlmEdge}.
Divide-and-conquer dispatch extends this local reasoning to mission
structure. Regions, roles, objectives, knowledge sources, model
capabilities, and edge-cloud placement can all serve as partition
criteria, but the local semantic decisions must eventually be
recombined into a physically consistent fleet plan
\cite{luo2025multiLlmEdge}.

This comparison suggests a practical boundary for LLM use in UAV dispatch.
The LLM layer is most defensible when it remains above direct flight control, where it can interpret missions, decompose tasks, orchestrate solvers, critique semantic consistency, and explain decisions to operators. Low-level actions still need retrieval grounding, tool constraints, simulation,
and runtime monitoring.

\subsection{Open Challenges}
\label{subsec:llm_open_challenges}

A failure can enter the dispatch process before any trajectory or schedule is generated.
An LLM may invent a constraint, misread a map description, ignore a battery or weather limit, or produce a plan that sounds coherent but cannot be executed.
In UAV missions, such errors are especially damaging because semantic interpretation affects task allocation, route generation, communication scheduling, and operator approval.
Retrieval-augmented generation reduces this risk
only when the retrieved context is relevant, current, and faithfully
used by the model
\cite{ru2024ragchecker,wu2026relationHallucinationTPAMI,yang2026defyingDistractionsTPAMI,wang2026vlBiasBenchTPAMI}.

Retrieved knowledge also has to survive the physical layer. LLMs do
not model flight dynamics, propulsion energy, wind disturbance,
sensing quality, channel variation, or collision avoidance with the
precision required by UAV operation. Their outputs should therefore be
treated as structured candidates for downstream motion planners,
communication models, digital twins, and runtime monitors. The
difficult part is not merely adding these checks, but making the
semantic-to-physical interface fast enough for low-altitude missions
where the environment changes during execution.

In the field, model choice is rarely independent of connectivity,
energy, and mission risk. Cloud LLMs offer stronger reasoning, but
they may be unavailable or too slow in remote inspection, emergency
response, and contested communication environments. Onboard models
reduce latency and protect sensitive data, yet they are smaller, more
energy constrained, and more brittle under complex instructions. Model
compression, split inference, edge-cloud fallback, and task-aware model
selection should therefore be part of dispatch policy itself. Once
multiple UAVs participate, the same resource problem appears as a
coordination problem: LLM agents can make swarm behavior easier to
explain, but unconstrained dialogue can also introduce verbose
negotiation, inconsistent beliefs, and hidden prompt dependencies.
Scalable multi-UAV use will require structured message schemas,
role-based communication, consensus rules, and local validators.

The same integration also widens the attack and evaluation surface.
The LLM layer may touch mission descriptions, locations, images,
videos, communication logs, and infrastructure data, so prompt
injection, data leakage, model inversion, unsafe tool use, and
ambiguous responsibility boundaries have to be addressed together with
access control, retrieval filtering, privacy-preserving inference, and
tamper-resistant logs. Evaluation also has to move beyond mission
time, energy, coverage, throughput, and path length. A benchmark for
LLM-enabled dispatch should test semantic correctness, constraint
satisfaction, plan executability, explanation quality, robustness to
prompt attacks, sim-to-real transfer, and the burden placed on human
approval. A plan that saves energy while violating a retrieved
airspace rule should be judged as a dispatch failure rather than as an
efficient solution.

\subsection{Future Research Directions}
\label{subsec:llm_hotspots}

The same interface view motivates the research directions summarized in Table~\ref{tab:llm_future_directions}, where semantic reasoning meets verified dispatch behavior.

\begin{table*}[t]
\centering
\caption{Future research directions for LLM-enabled UAV dispatch.}
\label{tab:llm_future_directions}
\footnotesize
\setlength{\arrayrulewidth}{0.4pt}
\setlength{\tabcolsep}{5pt}
\renewcommand{\arraystretch}{1.18}
\begin{tabular}{|>{\arraybackslash}m{0.20\textwidth}
                |>{\arraybackslash}m{0.34\textwidth}
                |>{\arraybackslash}m{0.36\textwidth}|}
\hline
\textbf{Hotspot} & \textbf{Dispatch Focus} & \textbf{Key Research Issues} \\
\hline
RAG-grounded dispatch knowledge bases &
Retrieve maps, regulations, weather, mission logs, and communication context for planning &
Source freshness, retrieval faithfulness, traceable constraint generation \\
\hline

LLM-tool-solver co-design &
Connect LLM reasoning with optimizers, simulators, planners, and safety monitors &
Stable interfaces, feasibility checking, repairable plan representations \\
\hline

Decentralized agentic dispatch &
Support local semantic interpretation, task claiming, peer negotiation, and replanning &
Latency, compact protocols, prompt safety, distributed verification \\
\hline

Semantic divide-and-conquer &
Partition missions by regions, roles, objectives, knowledge sources, and model placement &
Boundary conflicts, load balancing, cross-agent consistency \\
\hline

Multimodal aerial reasoning &
Fuse aerial images, videos, maps, text instructions, and sensor summaries &
Scene grounding, hallucination control, sim-to-real transfer \\
\hline

Edge multi-LLM orchestration &
Assign reasoning tasks across onboard, edge, and cloud models &
Model selection, fallback policies, energy and bandwidth limits \\
\hline

Safety-certified human governance &
Expose uncertainty, cite evidence, request approval, and log decisions &
Accountability, HITL design, auditability, regulatory compliance \\
\hline
\end{tabular}
\end{table*}

Future dispatch systems will need knowledge layers that combine maps,
airspace rules, weather feeds, maintenance records, mission logs, and
communication-state summaries. Such corpora are useful only if their
content can be traced back to source material and transformed into
constraints that solvers and monitors can check. Retrieval-augmented
generation therefore has to be studied together with
LLM-tool-solver co-design. A model may propose and revise a plan, but
optimizers, simulators, route planners, communication models, and
safety monitors must decide whether that plan remains feasible.

In missions such as inspection, environmental monitoring, search and
rescue, and post-disaster assessment, dispatch often depends on what
the platform sees, not only on what the operator requests. Damaged
infrastructure, blocked roads, moving targets, smoke, terrain changes,
and temporary crowding can all change task priority. MLLMs can connect
images, videos, maps, language instructions, and sensor streams, but
their outputs must be grounded against geometry, time, and mission
context. The model placement problem then becomes unavoidable. A small
onboard model may summarize local evidence, an edge model may
coordinate a nearby group of UAVs, and a cloud model may support
broader planning when connectivity is available. Dispatch must decide
when to invoke each level and how to recover when a model, link, or
sensor stream fails.

For large fleets, local reaction and global recomposition have to be studied together.
Local agents need compact negotiation, prompt-safe replanning, and distributed verification; decomposed missions need reliable boundaries for regions, roles, objectives, knowledge sources, and model placement.
Separating decentralized agentic dispatch from semantic divide-and-conquer dispatch is useful for taxonomy, but real systems will often mix them: a UAV may claim a nearby task, exchange a short justification
with neighbors, and then rely on a higher-level decomposition policy
to maintain load balance and mission consistency.

As these systems move toward deployment, governance can no longer be
added after planning has finished. LLM-enabled dispatch should expose
uncertainty, cite retrieved evidence, present alternatives, and request
human approval when safety, privacy, or regulation is at stake.
The architecture then remains hybrid in a deliberate way: the semantic layer interprets intent and organizes knowledge, while the symbolic and physical layers optimize assignments, trajectories, schedules, resources, and associations under explicit constraints.
The long-term value of LLMs in UAV dispatch will come from grounded, verifiable, privacy-aware, and human-governed interfaces, not from replacing dispatch algorithms with unconstrained text generation.

\section{Conclusion}
\label{sec:conclusion}

This survey has reviewed UAV dispatch in the LLM era from a system-level perspective.
We first established a unified view of UAV dispatch by connecting general dispatch models, unmanned-system dispatch, and UAV dispatch under a common set of decisions, including task assignment, trajectory planning, scheduling, resource allocation, and cooperation.
Based on how mission intent is transformed into executable fleet-level decisions, we organized existing studies into four representative paradigms: pipeline dispatch, global assignment dispatch, decentralized agentic dispatch, and divide-and-conquer dispatch.
These paradigms capture different dispatch logics:
pipeline dispatch emphasizes staged mission processing, 
global assignment dispatch exploits centralized state for joint optimization, 
decentralized agentic dispatch supports local autonomy and decentralized coordination, and divide-and-conquer dispatch reduces large-scale coupling through spatial, temporal, functional, role-based, knowledge-source, or model-placement decomposition.
Across these paradigms, recent studies have advanced UAV dispatch in trajectory planning, routing, task scheduling, energy management, communication optimization, privacy protection, semantic orchestration, and early LLM-agent coordination.

The emergence of LLMs and multimodal foundation models introduces a new semantic layer for UAV dispatch.
LLMs can interpret natural-language missions, retrieve contextual knowledge, decompose high-level objectives, orchestrate external solvers, support multi-agent negotiation, explain dispatch decisions, and assist human-in-the-loop revision.
However, they should not be viewed as replacements for optimization, control, communication theory, or safety verification.
In safety-critical low-altitude operations, language-based reasoning must be grounded in reliable evidence, transformed into structured dispatch candidates, and verified by specialized models, simulators, controllers, and runtime monitors.
Future UAV dispatch systems should therefore move toward hybrid semantic--symbolic--physical architectures that integrate LLM-enabled reasoning with mathematical optimization, multi-agent learning, privacy-preserving computation, digital twins, and formally verified safety mechanisms.
Such architectures will be essential for building scalable, trustworthy, and human-governed UAV dispatch systems in complex low-altitude environments.

\bibliographystyle{IEEEtran}
\bibliography{references}

@inproceedings{manzini2026deploying,
  title={Deploying Rapid Damage Assessments from sUAS Imagery for Disaster Response},
  author={Manzini, Thomas and Perali, Priyankari and Murphy, Robin R},
  booktitle={Proceedings of the AAAI Conference on Artificial Intelligence},
  volume={40},
  number={47},
  pages={40336--40343},
  year={2026},
  doi={10.1609/aaai.v40i47.41474},
}

@article{tagle2025effective,
  title={Effective integration of drone technology for mapping and managing palm species in the Peruvian Amazon},
  author={Tagle Casapia, Ximena and Cardenas-Vigo, Rodolfo and Marcos, Diego and et. al.},
  journal={Nat. Commun.},
  volume={16},
  number={1},
  pages={3764},
  year={2025},
  publisher={Nature Publishing Group UK London},
  doi={10.1038/s41467-025-58358-5},
}

@ARTICLE{11426817,
  author={Lyu, Wenjun and Wang, Haotian and Hong, Zhiqing and et. al.},
  journal={IEEE Trans. Mob. Comput.}, 
  title={Mobility-Aware Order Relay in Last-Mile Delivery}, 
  year={2026},
  volume={},
  number={},
  pages={1-15},
  doi={10.1109/TMC.2026.3671756}}

@article{zhu2024,
  title={Collaborative Reinforcement Learning Based Unmanned Aerial Vehicle (UAV) Trajectory Design for 3D UAV Tracking},
  author={Zhu, Y. and Chen, M. and Wang, S. and et. al.},
  journal={ IEEE Trans. Mob. Comput.},
  volume={23},
  number={12},
  pages={10787-10802},
  year={2024},
  doi={10.1109/tmc.2024.3382913},
}

@article{wang2022,
  title={Joint Optimization of UAV Trajectory and Sensor Uploading Powers for UAV-Assisted Data Collection in Wireless Sensor Networks},
  author={Wang, Y. and Chen, M. and Pan, C. and et. al.},
  journal={IEEE Internet Things},
  volume={9},
  number={13},
  pages={11214-11226},
  year={2022},
  doi={10.1109/jiot.2021.3126329},
}

@article{yanmaz2024,
  title={Dynamic Multi-UAV Path Planning for Multi-Target Search and Connectivity},
  author={Yanmaz, E. and Balanji, H. and G{\"u}ven, {\.I}.},
  journal={ IEEE Trans. Veh. Technol.},
  volume={73},
  number={7},
  pages={10516-10528},
  year={2024},
  doi={10.1109/tvt.2024.3363840},
}

@article{ren2025,
  title={Rolling Dispatch for AAVs Inspection Based on Task Adaptive Clustering},
  author={Ren, J. and Song, N. and Li, X. and et. al.},
  journal={IEEE T. Autom. Sci. Eng.},
  volume={22},
  pages={14108-14119},
  year={2025},
  doi={10.1109/tase.2025.3553745},
}

@ARTICLE{luan2022hierarchical,
  author={Luan, Qiuji and Cui, Hongyan and Zhang, Lifeng and et. al.},
  journal={IEEE Internet Things J.},
  title={A Hierarchical Hybrid Subtask Scheduling Algorithm in UAV-Assisted MEC Emergency Network},
  year={2022},
  volume={9},
  number={16},
  pages={14586-14599},
  doi={10.1109/JIOT.2021.3138263}
}

@ARTICLE{2025Mission,
  author={Cheng, Yujie and Ding, Yu and Lu, Chen},
  journal={Aerospace},
  title={Mission Planning for UAV Swarm with Aircraft Carrier Delivery: A Decoupled Framework},
  year={2025},
  volume={12},
  number={8},
  pages={691},
  doi={10.3390/aerospace12080691}
}

@ARTICLE{guo2025hierarchical,
  author={Guo, Anxin and Zhang, Zhenxing and Wu, Ao and et. al.},
  journal={Sensors},
  title={A Hierarchical Framework and Marginal Return Optimization for Dynamic Task Allocation in Heterogeneous UAV Networks},
  year={2025},
  volume={25},
  number={21},
  pages={6676},
  publisher={MDPI},
  doi={10.3390/s25216676}
}

@ARTICLE{zhao2026multi,
  author={Zhao, Xun and Liu, Yunping and Zhu, Shuaihui and et. al.},
  journal={J. Cloud Comput.-Adv. Syst. Appl.},
  title={Multi-stage task allocation strategy for UAV cluster in multi-object tracking},
  year={2026},
  volume={15},
  number={1},
  pages={21},
  publisher={Springer},
  doi={10.1186/s13677-025-00832-y}
}

@INPROCEEDINGS{junger2021talking,
  author={J{\"u}nger, Franz and Schopferer, Simon and Benders, Sebastian and et. al.},
  booktitle={2021 International Conference on Unmanned Aircraft Systems (ICUAS)},
  title={Talking to Autonomous Drones: Command and Control Based on Hierarchical Task Decomposition},
  year={2021},
  pages={968-977},
  organization={IEEE},
  doi={10.1109/ICUAS51884.2021.9476753}
}

@ARTICLE{li2025hierarchical,
  author={Li, Yiyuan and Chen, Weiyi and Fu, Bing and et. al.},
  journal={Drones},
  title={A Hierarchical Decoupling Task Planning Method for Multi-UAV Collaborative Multi-Region Coverage with Task Priority Awareness},
  year={2025},
  volume={9},
  number={8},
  pages={575},
  publisher={MDPI},
  doi={10.3390/drones9080575}
}

@ARTICLE{zhou2025hierarchical,
  author={Zhou, Zhenlin and Long, Teng and Sun, Jingliang and et. al.},
  journal={J. Syst. Eng. Electron.},
  title={Hierarchical cooperative path planning method using three-dimensional velocity-obstacle strategy for multiple fixed-wing UAVs},
  year={2025},
  volume={36},
  number={5},
  pages={1342-1352},
  publisher={BIAI},
  doi={10.23919/JSEE.2025.000087}
}

@article{liu2022,
  title={UAV Trajectory Optimization for Time-Constrained Data Collection in UAV-Enabled Environmental Monitoring Systems},
  author={Liu, K. and Zheng, J.},
  journal={IEEE Internet Things J.},
  volume={9},
  number={23},
  pages={24300-24314},
  year={2022},
  doi={10.1109/jiot.2022.3189214},
}

@article{wu2023,
  title={On the Interplay Between Sensing and Communications for UAV Trajectory Design},
  author={Wu, J. and Yuan, W. and Bai, L.},
  journal={IEEE Internet Things J.},
  volume={10},
  number={23},
  pages={20383-20395},
  year={2023},
  doi={10.1109/jiot.2023.3287991},
}

@article{zhu2021,
  title={UAV Trajectory Planning in Wireless Sensor Networks for Energy Consumption Minimization by Deep Reinforcement Learning},
  author={Zhu, B. and Bedeer, E. and Nguyen, H. and et. al.},
  journal={ IEEE Trans. Veh. Technol.},
  volume={70},
  number={9},
  pages={9540-9554},
  year={2021},
  doi={10.1109/tvt.2021.3102161},
}

@article{moon2021,
  title={Deep Reinforcement Learning Multi-UAV Trajectory Control for Target Tracking},
  author={Moon, J. and Papaioannou, S. and Laoudias, C. and et. al.},
  journal={IEEE Internet Things J.},
  volume={8},
  number={20},
  pages={15441-15455},
  year={2021},
  doi={10.1109/jiot.2021.3073973},
}

@article{he2023,
  title={Fairness-Based 3-D Multi-UAV Trajectory Optimization in Multi-UAV-Assisted MEC System},
  author={He, Y. and Gan, Y. and Cui, H. and et. al.},
  journal={IEEE Internet Things J.},
  volume={10},
  number={13},
  pages={11383-11395},
  year={2023},
  doi={10.1109/jiot.2023.3241087},
}

@article{kang2022,
  title={Improving Dual-UAV Aided Ground-UAV Bi-Directional Communication Security: Joint UAV Trajectory and Transmit Power Optimization},
  author={Kang, H. and Chang, X. and Misic, J. and et. al.},
  journal={IEEE Trans. Veh. Technol.},
  volume={71},
  number={10},
  pages={10570-10583},
  year={2022},
  doi={10.1109/tvt.2022.3184804},
}

@article{zhang2024isac,
  title={A Joint UAV Trajectory, User Association, and Beamforming Design Strategy for Multi-UAV-Assisted ISAC Systems},
  author={Zhang, R. and Zhang, Y. and Tang, R. and et. al.},
  journal={IEEE Internet Things J.},
  volume={11},
  number={18},
  pages={29360-29374},
  year={2024},
  doi={10.1109/jiot.2024.3430390},
}

@article{zhang2024secureRelay,
  title={UAV Swarm-Enabled Collaborative Secure Relay Communications With Time-Domain Colluding Eavesdropper},
  author={Zhang, C. and Sun, G. and Wu, Q. and et. al.},
  journal={IEEE Trans. Mob. Comput.},
  volume={23},
  number={9},
  pages={8601-8619},
  year={2024},
  doi={10.1109/tmc.2024.3350885},
}

@article{xiong2025,
  title={Collaborative coverage path planning for UAV swarm for multi-region post-disaster assessment},
  author={Xiong, Y. and Zhou, Y. and She, J. and et. al.},
  journal={Vehicular Communications},
  volume={53},
  pages={100915},
  year={2025},
  doi={10.1016/j.vehcom.2025.100915},
}

@article{mingnd,
  title={Comparative Analysis of Different UAV Swarm Control Methods on Unmanned Farms},
  author={Ming, R. and Jiang, R. and Luo, H. and et. al.},
  journal={Agronomy},
  volume={13},
  number={10},
  pages={2499},
  year={2023},
  doi={10.3390/agronomy13102499},
}

@article{hu2024,
  title={Computation-Efficient Grouping, Trajectory, and Resource Allocation for UAV Swarm-Assisted Aerial–Ground Collaborative Computing Networks},
  author={Hu, H. and Chen, Z. and Zhou, F. and et. al.},
  journal={IEEE Internet Things J.},
  volume={11},
  number={7},
  pages={12510-12525},
  year={2024},
  doi={10.1109/jiot.2023.3337840},
}

@article{dangngoc2022,
  title={Secure Swarm UAV-Assisted Communications With Cooperative Friendly Jamming},
  author={Dang-Ngoc, H. and Nguyen, D. and Ho-Van, K. and et. al.},
  journal={IEEE Internet Things J.},
  volume={9},
  number={24},
  pages={25596-25611},
  year={2022},
  doi={10.1109/jiot.2022.3197975},
}

@article{wu2021,
  title={Swarm-Based 4D Path Planning For Drone Operations in Urban Environments},
  author={Wu, Y. and Low, K. and Pang, B. and et. al.},
  journal={IEEE Trans. Veh. Technol.},
  volume={70},
  number={8},
  pages={7464-7479},
  year={2021},
  doi={10.1109/tvt.2021.3093318},
}

@article{alyassi2023,
  title={Autonomous Recharging and Flight Mission Planning for Battery-Operated Autonomous Drones},
  author={Alyassi, R. and Khonji, M. and Karapetyan, A. and et. al.},
  journal={IEEE Trans. Autom. Sci. Eng.},
  volume={20},
  number={2},
  pages={1034-1046},
  year={2023},
  doi={10.1109/tase.2022.3175565},
}

@article{shi2021,
  title={Drone-Cell Trajectory Planning and Resource Allocation for Highly Mobile Networks: A Hierarchical DRL Approach},
  author={Shi, W. and Li, J. and Wu, H. and et. al.},
  journal={IEEE Internet Things J.},
  volume={8},
  number={12},
  pages={9800-9813},
  year={2021},
  doi={10.1109/jiot.2020.3020067},
}

@article{montemannind,
  title={Solving the Parallel Drone Scheduling Traveling Salesman Problem via Constraint Programming},
  author={Montemanni, R. and Dell’Amico, M.},
  journal={Algorithms},
  volume={16},
  number={1},
  pages={40},
  year={2023},
  doi={10.3390/a16010040},
}

@article{campuzano2022,
  title={The Dynamic Drone Scheduling Delivery Problem},
  author={Campuzano, G. and Lalla-Ruiz, E. and Mes, M.},
  journal={LNCS},
  pages={260-274},
  year={2022},
  doi={10.1007/978-3-031-16579-5_18},
}

@article{rigas2022,
  title={Scheduling a Fleet of Drones for Monitoring Missions With Spatial, Temporal, and Energy Constraints},
  author={Rigas, E. and Kolios, P. and Mavrovouniotis, M. and et. al.},
  journal={IEEE Trans. Intell. Transp. Syst.},
  volume={23},
  number={9},
  pages={15133-15145},
  year={2022},
  doi={10.1109/tits.2021.3137359},
}

@article{khochare2024,
  title={Improved Algorithms for Co-Scheduling of Edge Analytics and Routes for UAV Fleet Missions},
  author={Khochare, A. and Sorbelli, F. and Simmhan, Y. and et. al.},
  journal={IEEE ACM T NETWORK},
  volume={32},
  number={1},
  pages={17-33},
  year={2024},
  doi={10.1109/tnet.2023.3277810},
}

@article{zaind,
  title={A Drone Scheduling Method for Emergency Power Material Transportation Based on Deep Reinforcement Learning Optimized PSO Algorithm},
  author={Zai, W. and Wang, J. and Li, G.},
  journal={Sustainability},
  volume={15},
  number={17},
  pages={13127},
  year={2023},
  doi={10.3390/su151713127},
}

@article{yang2025,
  title={A bi-objective optimisation model for the drone scheduling problem in island delivery},
  author={Yang, Y. and Liu, J. and Wang, S.},
  journal={Int. J. Prod. Res.},
  volume={63},
  number={19},
  pages={7174-7195},
  year={2025},
  doi={10.1080/00207543.2025.2496965},
}

@article{khochare2021,
  title={Heuristic Algorithms for Co-scheduling of Edge Analytics and Routes for UAV Fleet Missions},
  author={Khochare, A. and Simmhan, Y. and Sorbelli, F. and et. al.},
  journal={INFOCOM},
  pages={1-10},
  year={2021},
  doi={10.1109/infocom42981.2021.9488740},
}

@article{gongnd,
  title={Research on the Design of a Priority-Based Multi-Stage Emergency Material Scheduling System for Drone Coordination},
  author={Gong, S. and Chen, G. and Yang, Z.},
  journal={Drones},
  volume={9},
  number={8},
  pages={524},
  year={2025},
  doi={10.3390/drones9080524},
}

@article{torabbeigi2021,
  title={An Optimization Approach to Minimize the Expected Loss of Demand Considering Drone Failures in Drone Delivery Scheduling},
  author={Torabbeigi, M. and Lim, G. and Ahmadian, N. and et. al.},
  journal={J. Intell. Robot. Syst.},
  volume={102},
  number={1},
  year={2021},
  doi={10.1007/s10846-021-01370-w},
}

@article{pachayappan2021,
  title={A Solution to Drone Routing Problems using Docking Stations for Pickup and Delivery Services},
  author={Pachayappan, M. and Sudhakar, V.},
  journal={Transp. Res. Rec.},
  volume={2675},
  number={12},
  pages={1056-1074},
  year={2021},
  doi={10.1177/03611981211032219},
}

@article{zhao2023,
  title={UAV Dispatch Planning for a Wireless Rechargeable Sensor Network for Bridge Monitoring},
  author={Zhao, C. and Wang, Y. and Zhang, X. and et. al.},
  journal={IEEE Trans. Sustainable Comput.},
  volume={8},
  number={2},
  pages={293-309},
  year={2023},
  doi={10.1109/tsusc.2022.3224442},
}

@article{rinaldind,
  title={Urban Air Logistics with Unmanned Aerial Vehicles (UAVs): Double-Chromosome Genetic Task Scheduling with Safe Route Planning},
  author={Rinaldi, M. and Primatesta, S. and Bugaj, M. and et. al.},
  journal={Smart Cities},
  volume={7},
  number={5},
  pages={2842-2860},
  year={2024},
  doi={10.3390/smartcities7050110},
}

@article{oubbati2021,
  title={Dispatch of UAVs for Urban Vehicular Networks: A Deep Reinforcement Learning Approach},
  author={Oubbati, O. and Atiquzzaman, M. and Baz, A. and et. al.},
  journal={IEEE Trans. Veh. Technol.},
  volume={70},
  number={12},
  pages={13174-13189},
  year={2021},
  doi={10.1109/tvt.2021.3119070},
}

@article{liundInspection,
  title={Advancing Multi-UAV Inspection Dispatch Based on Bilevel Optimization and GA-NSGA-II},
  author={Liu, Y. and Chen, C. and Sun, Y. and et. al.},
  journal={Applied Sciences},
  volume={15},
  number={7},
  pages={3673},
  year={2025},
  doi={10.3390/app15073673},
}

@article{bettisorbelli2025,
  title={Dispatching point selection for a drone-based delivery system operating in a mixed Euclidean–Manhattan grid},
  author={Betti Sorbelli, F. and Corò, F. and Das, S. and et. al.},
  journal={Annals of Operations Research},
  volume={351},
  number={1},
  pages={203-222},
  year={2025},
  doi={10.1007/s10479-023-05458-4},
}

@article{pei2023,
  title={Urban On-Demand Delivery via Autonomous Aerial Mobility: Formulation and Exact Algorithm},
  author={Pei, Z. and Fang, T. and Weng, K. and Yi, W.},
  journal={IEEE TASE},
  volume={20},
  number={3},
  pages={1675-1689},
  year={2023},
  doi={10.1109/tase.2022.3184324},
}

@article{joo2026,
  title={A Branch-and-Price Algorithm for Robust Drone-Vehicle Routing Problem with Time Windows},
  author={Joo, J. and Lee, C.},
  journal={INFORMS J Comput},
  volume={38},
  number={1},
  pages={102-125},
  year={2026},
  doi={10.1287/ijoc.2023.0484},
}

@article{mao2025,
  title={DL-DRL: A Double-Level Deep Reinforcement Learning Approach for Large-Scale Task Scheduling of Multi-UAV},
  author={Mao, X. and Wu, G. and Fan, M. and et. al.},
  journal={IEEE Trans. Autom. Sci. Eng.},
  volume={22},
  pages={1028-1044},
  year={2025},
  doi={10.1109/tase.2024.3358894},
}

@article{akram2025,
  title={Privacy-Preserving Spatial Crowdsourcing Drone Services for Postdisaster Infrastructure Monitoring: A Conditional Federated Learning Approach},
  author={Akram, J. and Akram, A. and Ingle, P. and et. al.},
  journal={IEEE J-STARS},
  volume={18},
  pages={16272-16291},
  year={2025},
  doi={10.1109/jstars.2025.3577648},
}

@article{pu2023,
  title={Privacy-Preserving and Fault-Tolerant Data Aggregation Protocol for Internet of Drones},
  author={Pu, C.},
  journal={DSC},
  pages={1-8},
  year={2023},
  doi={10.1109/dsc61021.2023.10354242},
}

@article{huang2024,
  title={Privacy-Aware Design and Analysis of Drone Remote Identification Systems},
  author={Huang, H. and Fang, Y. and Mazotti, B. and et. al.},
  journal={IEEE Trans. Intell. Transp. Syst.},
  volume={25},
  number={12},
  pages={21455-21468},
  year={2024},
  doi={10.1109/tits.2024.3456032},
}

@article{jagarlapudi2022,
  title={Drone Helps Privacy: Sky Caching Assisted $k$-Anonymity in Spatial Querying},
  author={Jagarlapudi, H. and Lim, S. and Chae, J. and et. al.},
  journal={IEEE Systems Journal},
  volume={16},
  number={4},
  pages={6360-6370},
  year={2022},
  doi={10.1109/jsyst.2022.3171211},
}

@article{ebrahimi2021,
  title={Autonomous UAV Trajectory for Localizing Ground Objects: A Reinforcement Learning Approach},
  author={Ebrahimi, D. and Sharafeddine, S. and Ho, P. and et. al.},
  journal={IEEE Trans. Mob. Comput.},
  volume={20},
  number={4},
  pages={1312-1324},
  year={2021},
  doi={10.1109/tmc.2020.2966989},
}

@article{nguyen2022,
  title={3D UAV Trajectory and Data Collection Optimisation Via Deep Reinforcement Learning},
  author={Nguyen, K. and Duong, T. and Do-Duy, T. and et. al.},
  journal={IEEE Trans. Wireless Commun.},
  volume={70},
  number={4},
  pages={2358-2371},
  year={2022},
  doi={10.1109/tcomm.2022.3148364},
}

@article{shukla2025,
  title={Trajectory-Prediction Techniques for Unmanned Aerial Vehicles (UAVs): A Comprehensive Survey},
  author={Shukla, P. and Shukla, S. and Kumar Singh, A.},
  journal={IEEE Commun Surv Tut},
  volume={27},
  number={3},
  pages={1867-1910},
  year={2025},
  doi={10.1109/comst.2024.3471671},
}

@article{zhu2023,
  title={UAV Trajectory Planning for AoI-Minimal Data Collection in UAV-Aided IoT Networks by Transformer},
  author={Zhu, B. and Bedeer, E. and Nguyen, H. and et. al.},
  journal={IEEE Trans. Wireless Commun.},
  volume={22},
  number={2},
  pages={1343-1358},
  year={2023},
  doi={10.1109/twc.2022.3204438},
}

@article{zhao2021,
  title={Multi-UAV Trajectory Planning for Energy-Efficient Content Coverage: A Decentralized Learning-Based Approach},
  author={Zhao, C. and Liu, J. and Sheng, M. and et. al.},
  journal={IEEE J Sel Area Comm},
  volume={39},
  number={10},
  pages={3193-3207},
  year={2021},
  doi={10.1109/jsac.2021.3088669},
}

@article{huang2022,
  title={Online UAV Trajectory Planning for Covert Video Surveillance of Mobile Targets},
  author={Huang, H. and Savkin, A. and Ni, W.},
  journal={IEEE Trans. Autom. Sci. Eng.},
  volume={19},
  number={2},
  pages={735-746},
  year={2022},
  doi={10.1109/tase.2021.3062810},
}

@article{xu2022,
  title={Autonomous Obstacle Avoidance and Target Tracking of UAV Based on Deep Reinforcement Learning},
  author={Xu, G. and Jiang, W. and Wang, Z. and et. al.},
  journal={J. Intell. Robot. Syst.},
  volume={104},
  number={4},
  year={2022},
  doi={10.1007/s10846-022-01601-8},
}

@article{huang2025,
  title={Biomimetic Multi-UAV Swarm Exploration With U2U Communications Under Resource Constraints},
  author={Huang, Y. and Wang, H. and Bai, X. and et. al.},
  journal={IEEE Trans. Veh. Technol.},
  volume={74},
  number={6},
  pages={9750-9766},
  year={2025},
  doi={10.1109/tvt.2025.3541299},
}

@article{hasan2023,
  title={3D relative directions based evolutionary computation for UAV-to-UAV interaction in swarm intelligence enabled decentralized networks},
  author={Hasan, M. and Rayhan Kabir, S. and Abdullah, S. and et. al.},
  journal={Alex. Eng. J.},
  volume={85},
  pages={104-113},
  year={2023},
  doi={10.1016/j.aej.2023.10.063},
}

@article{zhang2023,
  title={Game of Drones: Multi-UAV Pursuit-Evasion Game With Online Motion Planning by Deep Reinforcement Learning},
  author={Zhang, R. and Zong, Q. and Zhang, X. and et. al.},
  journal={IEEE Trans. Neural Netw. Learn. Syst.},
  volume={34},
  number={10},
  pages={7900-7909},
  year={2023},
  doi={10.1109/tnnls.2022.3146976},
}

@article{chinthi-reddy2022,
  title={DarkSky: Privacy-preserving target tracking strategies using a flying drone},
  author={Chinthi-Reddy, S. and Lim, S. and Choi, G. and et. al.},
  journal={Vehicular Communications},
  volume={35},
  pages={100459},
  year={2022},
  doi={10.1016/j.vehcom.2022.100459},
}

@article{sciancalepore2022,
  title={Privacy-Preserving Trajectory Matching on Autonomous Unmanned Aerial Vehicles},
  author={Sciancalepore, S. and George, D.},
  journal={ACSAC '22},
  pages={1-12},
  year={2022},
  doi={10.1145/3564625.3564626},
}

@article{tedeschi2023,
  title={PPCA - Privacy-Preserving Collision Avoidance for Autonomous Unmanned Aerial Vehicles},
  author={Tedeschi, P. and Sciancalepore, S. and Di Pietro, R.},
  journal={IEEE T Depend Secure},
  volume={20},
  number={2},
  pages={1541-1558},
  year={2023},
  doi={10.1109/tdsc.2022.3159837},
}

@article{gaond,
  title={UAV-Centric Privacy-Preserving Computation Offloading in Multi-UAV Mobile Edge Computing},
  author={Gao, C. and Wei, D. and Li, K. and et. al.},
  journal={Drones},
  volume={9},
  number={10},
  pages={701},
  year={2025},
  doi={10.3390/drones9100701},
}

@article{venturini2020distributed,
  author={Venturini, Federico and Mason, Federico and Pase, Francesco and et. al.},
  title={Distributed Reinforcement Learning for Flexible and Efficient UAV Swarm Control},
  journal={IEEE Trans. Cognit. Commun. Networking},
  volume={7},
  number={3},
  pages={955-969},
  year={2021},
  doi={10.1109/TCCN.2021.3063170},
}

@article{qin2025strategic,
  title={Strategic coordination of drones via short-term distributed optimization and long-term reinforcement learning},
  author={Qin, Chuhao and Pournaras, Evangelos},
  journal={Appl. Soft Comput.},
  volume={186},
  pages={114270},
  year={2026},
  publisher={Elsevier},                                               doi={10.1016/j.asoc.2025.114270},
}

@article{alkahtani2026explainable,
  title={Explainable multi agent reinforcement learning framework for secure and adaptive communication in UAV swarm based fanets},
  author={Alkahtani, Hend Khalid and Galiya, Ybytayeva and Akbayan, Bekarystankyzy and et. al.},
  journal={Scientific Reports},
  year={2026},
  publisher={Nature Publishing Group UK London}, 
  doi={10.1038/s41598-026-39366-x},
}

@article{javed2023,
  title={UAV trajectory planning for disaster scenarios},
  author={Javed, S. and Hassan, A. and Ahmad, R. and et. al.},
  journal={Vehicular Communications},
  volume={39},
  pages={100568},
  year={2023},
  doi={10.1016/j.vehcom.2022.100568},
}

@article{meng2022,
  title={UAV Trajectory and Beamforming Optimization for Integrated Periodic Sensing and Communication},
  author={Meng, K. and Wu, Q. and Ma, S. and et. al.},
  journal={IEEE Wireless Commun. Lett.},
  volume={11},
  number={6},
  pages={1211-1215},
  year={2022},
  doi={10.1109/lwc.2022.3161338},
}

@article{jing2024,
  title={ISAC From the Sky: UAV Trajectory Design for Joint Communication and Target Localization},
  author={Jing, X. and Liu, F. and Masouros, C. and et. al.},
  journal={IEEE Trans. Wireless Commun.},
  volume={23},
  number={10},
  pages={12857-12872},
  year={2024},
  doi={10.1109/twc.2024.3396571},
}

@article{liu202501,
  title={Flight path planning of UAV-driven refinement inspection for construction sites based on 3D reconstruction},
  author={Liu, X. and Yi, W. and Chen, P. and et. al.},
  journal={Autom. Constr.},
  volume={177},
  pages={106360},
  year={2025},
  doi={10.1016/j.autcon.2025.106360},
}

@article{cao2024,
  title={Computational Intelligence Algorithms for UAV Swarm Networking and Collaboration: A Comprehensive Survey and Future Directions},
  author={Cao, P. and Lei, L. and Cai, S. and et. al.},
  journal={IEEE COMMUN SURV TUT},
  volume={26},
  number={4},
  pages={2684-2728},
  year={2024},
  doi={10.1109/comst.2024.3395358},
}

@article{tang2023,
  title={Swarm intelligence algorithms for multiple unmanned aerial vehicles collaboration: a comprehensive review},
  author={Tang, J. and Duan, H. and Lao, S.},
  journal={ARTIF INTELL REV},
  volume={56},
  number={5},
  pages={4295-4327},
  year={2023},
  doi={10.1007/s10462-022-10281-7},
}

@article{ghambari2024,
  title={UAV path planning techniques: a survey},
  author={Ghambari, S. and Golabi, M. and Jourdan, L. and et. al.},
  journal={RAIRO-OPER RES},
  volume={58},
  number={4},
  pages={2951-2989},
  year={2024},
  doi={10.1051/ro/2024073},
}

@article{george2023,
  title={Privacy-Preserving Multi-Party Access Control for Third-Party UAV Services},
  author={George, D. and Sciancalepore, S. and Zannone, N.},
  journal={SACMAT 2023},
  pages={19-30},
  year={2023},
  doi={10.1145/3589608.3593837},
}

@article{enayati2023,
  title={Location Privacy Protection for UAVs in Package Delivery and IoT Data Collection},
  author={Enayati, S. and Goeckel, D. and Houmansadr, A. and et. al.},
  journal={IEEE Internet Things J. },
  volume={10},
  number={23},
  pages={20586-20601},
  year={2023},
  doi={10.1109/jiot.2023.3293755},
}

@article{sciancalepore2024,
  title={Privacy and Confidentiality Issues in Drone Operations: Challenges and Road Ahead},
  author={Sciancalepore, S.},
  journal={IEEE Network},
  volume={38},
  number={6},
  pages={227-233},
  year={2024},
  doi={10.1109/mnet.2024.3432730},
}

@article{Gong_2026,
  title={Safe and Economical UAV Trajectory Planning in Low-Altitude Airspace: A Hybrid DRL-LLM Algorithm With Compliance Awareness},
  author={Gong, Yanwei and Fan, Junchao and Zhang, Ruichen and et. al.},
  journal={IEEE Trans. Mob. Comput.},
  pages={1--17},
  year={2026},
  doi={10.1109/tmc.2026.3668209},
}

@article{Pongsakornsathien_2025,
  title={Advances in low-altitude airspace management for uncrewed aircraft and advanced air mobility},
  author={Pongsakornsathien, Nichakorn and Safwat, Nour El-Din and others},
  journal={Prog. Aerosp. Sci.},
  volume={154},
  pages={101085},
  year={2025},
  doi={10.1016/j.paerosci.2025.101085},
}

@article{Chang_2024,
  title={A Survey on Evaluation of Large Language Models},
  author={Chang, Yupeng and Wang, Xu and Wang, Jindong and et. al.},
  journal={ACM T INTEL SYST TEC},
  volume={15},
  number={3},
  pages={1--45},
  year={2024},
  doi={10.1145/3641289},
}

@article{li2025llmIsacUav,
  title={Large Language Model Based Multi-Objective Optimization for Integrated Sensing and Communications in UAV Networks},
  author={Li, Haoyun and Xiao, Ming and Wang, Kezhi and et. al.},
  journal={IEEE Wireless Commun. Lett.},
  volume={14},
  number={4},
  pages={979--983},
  year={2025},
  doi={10.1109/LWC.2025.3529082},
}

@article{luo2025multiLlmEdge,
  title={Toward Edge General Intelligence with Multiple-Large Language Model (Multi-LLM): Architecture, Trust, and Orchestration},
  author={Luo, Haoxiang and Liu, Yinqiu and Zhang, Ruichen and et. al.},
  journal={IEEE Trans. Cognit. Commun. Networking},
  volume={11},
  number={6},
  pages={3563--3585},
  year={2025},
}

@article{liu2025llmInspection,
  title={A LLM-Informed Multi-Agent AI System for Drone-Based Visual Inspection for Infrastructure},
  author={Liu, J. and Li, H. and Chai, C. and et. al.},
  journal={Adv. Eng. Inform.},
  volume={68},
  pages={103643},
  year={2025},
  doi={10.1016/j.aei.2025.103643},
}

@inproceedings{liu2023aerialvln,
  title={AerialVLN: Vision-and-Language Navigation for UAVs},
  author={Liu, Shubo and Zhang, Hongsheng and Qi, Yuankai and et. al.},
  booktitle={Proceedings of the IEEE/CVF International Conference on Computer Vision},
  pages={15384--15394},
  year={2023},
}

@inproceedings{wu2025aeroduo,
  title={AeroDuo: Aerial Duo for UAV-Based Vision and Language Navigation},
  author={Wu, Rui and Zhang, Yiran and Chen, Jie and et. al.},
  booktitle={Proceedings of the 33rd ACM International Conference on Multimedia},
  pages={2576--2585},
  year={2025},
}

@inproceedings{lewis2020rag,
  title={Retrieval-Augmented Generation for Knowledge-Intensive NLP Tasks},
  author={Lewis, Patrick and Perez, Ethan and Piktus, Aleksandra and et. al.},
  booktitle={Advances in Neural Information Processing Systems},
  volume={33},
  year={2020},
  url={https://proceedings.neurips.cc/paper/2020/hash/6b493230205f780e1bc26945df7481e5-Abstract.html},
}

@inproceedings{shen2023hugginggpt,
  title={HuggingGPT: Solving AI Tasks with ChatGPT and its Friends in Hugging Face},
  author={Shen, Yongliang and Song, Kaitao and Tan, Xu and et. al.},
  booktitle={Advances in Neural Information Processing Systems},
  volume={36},
  year={2023},
  url={https://proceedings.neurips.cc/paper_files/paper/2023/hash/77c33e6a367922d003ff102ffb92b658-Abstract-Conference.html},
}

@inproceedings{driess2023palmE,
  title={{PaLM}-E: An Embodied Multimodal Language Model},
  author={Driess, Danny and Xia, Fei and Sajjadi, Mehdi S. M. and et. al.},
  booktitle={Proceedings of the 40th International Conference on Machine Learning},
  series={Proceedings of Machine Learning Research},
  volume={202},
  pages={8469--8488},
  year={2023},
  url={https://proceedings.mlr.press/v202/driess23a.html},
}

@article{romera2024funsearch,
  title={Mathematical Discoveries from Program Search with Large Language Models},
  author={Romera-Paredes, Bernardino and Barekatain, Mohammadamin and Novikov, Alexander and et. al.},
  journal={Nature},
  volume={625},
  pages={468--475},
  year={2024},
  doi={10.1038/s41586-023-06924-6},
  url={https://www.nature.com/articles/s41586-023-06924-6},
}

@inproceedings{wei2022chainOfThought,
  title={Chain-of-Thought Prompting Elicits Reasoning in Large Language Models},
  author={Wei, Jason and Wang, Xuezhi and Schuurmans, Dale and et. al.},
  booktitle={Proceedings of the 36th International Conference on Neural Information Processing Systems},
  year={2022},
}

@article{madaan2023selfRefine,
  title={Self-Refine: Iterative Refinement with Self-Feedback},
  author={Madaan, Aman and Tandon, Niket and Gupta, Prakhar and et. al.},
  journal={NeurIPS},
  volume={36},
  pages={46534--46594},
  year={2023},
}

@article{ru2024ragchecker,
  title={RAGChecker: A Fine-Grained Framework for Diagnosing Retrieval-Augmented Generation},
  author={Ru, Dongyu and Qiu, Lin and Hu, Xiangkun and et. al.},
  journal={NeurIPS},
  volume={37},
  pages={21999--22027},
  year={2024},
}

@inproceedings{guo2024llmMultiAgentsSurvey,
  title={Large Language Model Based Multi-agents: A Survey of Progress and Challenges},
  author={Guo, Taicheng and Chen, Xiuying and Wang, Yaqi and et. al.},
  booktitle={Proceedings of the Thirty-Third International Joint Conference on Artificial Intelligence},
  pages={8048--8057},
  year={2024},
  note={Survey Track},
  doi={10.24963/ijcai.2024/890},
  url={https://doi.org/10.24963/ijcai.2024/890},
}

@inproceedings{yao2023react,
  title={ReAct: Synergizing Reasoning and Acting in Language Models},
  author={Yao, Shunyu and Zhao, Jeffrey and Yu, Dian and et. al.},
  booktitle={International Conference on Learning Representations},
  year={2023},
}

@article{shinn2023reflexion,
  title={Reflexion: Language Agents with Verbal Reinforcement Learning},
  author={Shinn, Noah and Cassano, Federico and Gopinath, Ashwin and et. al.},
  journal={NeurIPS},
  volume={36},
  pages={8634--8652},
  year={2023},
}

@article{han2023vitSurveyTPAMI,
  title={A Survey on Vision Transformer},
  author={Han, Kai and Wang, Yunhe and Chen, Hanting and et. al.},
  journal={IEEE TPAMI},
  volume={45},
  number={1},
  pages={87--110},
  year={2023},
  doi={10.1109/TPAMI.2022.3152247},
}

@article{zhang2024vlmSurveyTPAMI,
  title={Vision-Language Models for Vision Tasks: A Survey},
  author={Zhang, Jingyi and Huang, Jiaxing and Jin, Sheng and et. al.},
  journal={IEEE TPAMI},
  volume={46},
  number={8},
  pages={5625--5644},
  year={2024},
  doi={10.1109/TPAMI.2024.3369699},
}

@article{lin2024correctableLandmarkTPAMI,
  title={Correctable Landmark Discovery via Large Models for Vision-Language Navigation},
  author={Lin, Bingqian and Nie, Yunshuang and Wei, Ziming and et. al.},
  journal={IEEE TPAMI},
  volume={46},
  number={12},
  pages={8534--8548},
  year={2024},
  doi={10.1109/TPAMI.2024.3407759},
}

@article{xu2025lvlmEhubTPAMI,
  title={{LVLM-EHub}: A Comprehensive Evaluation Benchmark for Large Vision-Language Models},
  author={Xu, Peng and Shao, Wenqi and Zhang, Kaipeng and et. al.},
  journal={IEEE TPAMI},
  volume={47},
  number={3},
  pages={1877--1893},
  year={2025},
  doi={10.1109/TPAMI.2024.3507000},
}

@article{wu2026relationHallucinationTPAMI,
  title={Evaluating and Mitigating Relationship Hallucinations in Large Vision-Language Models},
  author={Wu, Mingrui and Li, Jiale and Ji, Jiayi and et. al.},
  journal={IEEE TPAMI},
  volume={48},
  number={6},
  pages={6332--6346},
  year={2026},
  doi={10.1109/TPAMI.2026.3656175},
}

@article{yang2026defyingDistractionsTPAMI,
  title={Defying Distractions in Multimodal Tasks: A Novel Benchmark for Large Vision-Language Models},
  author={Yang, Jinhui and Jiang, Ming and Zhao, Qi},
  journal={IEEE TPAMI},
  volume={48},
  number={6},
  pages={6314--6331},
  year={2026},
  doi={10.1109/TPAMI.2026.3655641},
}

@article{huang2025causalityMLLMTPAMI,
  title={A Causality-Aware Paradigm for Evaluating Creativity of Multimodal Large Language Models},
  author={Huang, Zhongzhan and Zhong, Shanshan and Zhou, Pan and et. al.},
  journal={IEEE TPAMI},
  volume={47},
  number={5},
  pages={3830--3846},
  year={2025},
  doi={10.1109/TPAMI.2025.3539433},
}

@article{zhang2026beyondLlavaHdTPAMI,
  title={Beyond {LLaVA-HD}: Diving Into High-Resolution Multimodal Large Language Models},
  author={Zhang, Yifan and Wen, Qingsong and Fu, Chaoyou and et. al.},
  journal={IEEE TPAMI},
  volume={48},
  number={5},
  pages={5493--5504},
  year={2026},
  doi={10.1109/TPAMI.2026.3650761},
}

@article{wang2026vlBiasBenchTPAMI,
  title={{VLBiasBench}: A Comprehensive Benchmark for Evaluating Bias in Large Vision-Language Model},
  author={Wang, Sibo and Cao, Xiangkui and Zhang, Jie and et. al.},
  journal={IEEE TPAMI},
  pages={1--14},
  year={2026},
  doi={10.1109/TPAMI.2026.3683747},
}

@article{wang2026imagineNavTPAMI,
  title={{ImagineNav++}: Prompting Vision-Language Models as Embodied Navigator Through Scene Imagination},
  author={Wang, Teng and Zhao, Xinxin and Cai, Wenzhe and et. al.},
  journal={IEEE TPAMI},
  pages={1--18},
  year={2026},
  doi={10.1109/TPAMI.2026.3688692},
}

@article{lin2026evolveNavTPAMI,
  title={{EvolveNav}: Empowering {LLM}-Based Vision-Language Navigation via Self-Improving Embodied Reasoning},
  author={Lin, Bingqian and Nie, Yunshuang and Zai, Khun Loun and et. al.},
  journal={IEEE TPAMI},
  pages={1--15},
  year={2026},
  doi={10.1109/TPAMI.2026.3682514},
}

@article{chen2025constraintAwareVLNTPAMI,
  title={Constraint-Aware Zero-Shot Vision-Language Navigation in Continuous Environments},
  author={Chen, Kehan and An, Dong and Huang, Yan and et. al.},
  journal={IEEE TPAMI},
  volume={47},
  number={11},
  pages={10441--10456},
  year={2025},
  doi={10.1109/TPAMI.2025.3594204},
}

@article{lin2025navcotTPAMI,
  title={{NavCoT}: Boosting {LLM}-Based Vision-and-Language Navigation via Learning Disentangled Reasoning},
  author={Lin, Bingqian and Nie, Yunshuang and Wei, Ziming and et. al.},
  journal={IEEE TPAMI},
  volume={47},
  number={7},
  pages={5945--5957},
  year={2025},
  doi={10.1109/TPAMI.2025.3554559},
}

@article{zhu2024openVocabularySurveyTPAMI,
  title={A Survey on Open-Vocabulary Detection and Segmentation: Past, Present, and Future},
  author={Zhu, Chaoyang and Chen, Long},
  journal={IEEE TPAMI},
  volume={46},
  number={12},
  pages={8954--8975},
  year={2024},
  doi={10.1109/TPAMI.2024.3413013},
}

@article{zhu2025llafsTPAMI,
  title={{LLaFS++}: Few-Shot Image Segmentation With Large Language Models},
  author={Zhu, Lanyun and Chen, Tianrun and Ji, Deyi and et. al.},
  journal={IEEE TPAMI},
  volume={47},
  number={9},
  pages={7715--7732},
  year={2025},
  doi={10.1109/TPAMI.2025.3573609},
}

@article{hu2023moMixTPAMI,
  title={{MO-MIX}: Multi-Objective Multi-Agent Cooperative Decision-Making With Deep Reinforcement Learning},
  author={Hu, Tianmeng and Luo, Biao and Yang, Chunhua and et. al.},
  journal={IEEE TPAMI},
  volume={45},
  number={10},
  pages={12098--12112},
  year={2023},
  doi={10.1109/TPAMI.2023.3283537},
}

@article{zhu2023transferDrlSurveyTPAMI,
  title={Transfer Learning in Deep Reinforcement Learning: A Survey},
  author={Zhu, Zhuangdi and Lin, Kaixiang and Jain, Anil K. and et. al.},
  journal={IEEE TPAMI},
  volume={45},
  number={11},
  pages={13344--13362},
  year={2023},
  doi={10.1109/TPAMI.2023.3292075},
}

@inproceedings{radford2021clip,
  title={Learning Transferable Visual Models From Natural Language Supervision},
  author={Radford, Alec and Kim, Jong Wook and Hallacy, Chris and et. al.},
  booktitle={Proceedings of the 38th International Conference on Machine Learning},
  series={Proceedings of Machine Learning Research},
  volume={139},
  pages={8748--8763},
  year={2021},
}

@inproceedings{alayrac2022flamingo,
  title={Flamingo: A Visual Language Model for Few-Shot Learning},
  author={Alayrac, Jean-Baptiste and Donahue, Jeff and Luc, Pauline and et. al.},
  booktitle={Advances in Neural Information Processing Systems},
  volume={35},
  pages={23716--23736},
  year={2022},
}

@inproceedings{li2023blip2,
  title={{BLIP-2}: Bootstrapping Language-Image Pre-training With Frozen Image Encoders and Large Language Models},
  author={Li, Junnan and Li, Dongxu and Savarese, Silvio and et. al.},
  booktitle={Proceedings of the 40th International Conference on Machine Learning},
  series={Proceedings of Machine Learning Research},
  volume={202},
  pages={19730--19742},
  year={2023},
}

@inproceedings{liu2023llava,
  title={Visual Instruction Tuning},
  author={Liu, Haotian and Li, Chunyuan and Wu, Qingyang and et. al.},
  booktitle={Advances in Neural Information Processing Systems},
  volume={36},
  year={2023},
}

@inproceedings{kirillov2023segmentAnything,
  title={Segment Anything},
  author={Kirillov, Alexander and Mintun, Eric and Ravi, Nikhila and et. al.},
  booktitle={Proceedings of the IEEE/CVF International Conference on Computer Vision},
  pages={4015--4026},
  year={2023},
}

@inproceedings{schick2023toolformer,
  title={Toolformer: Language Models Can Teach Themselves to Use Tools},
  author={Schick, Timo and Dwivedi-Yu, Jane and Dess{\`i}, Roberto and et. al.},
  booktitle={Advances in Neural Information Processing Systems},
  volume={36},
  year={2023},
}

@inproceedings{yao2023treeOfThoughts,
  title={Tree of Thoughts: Deliberate Problem Solving With Large Language Models},
  author={Yao, Shunyu and Yu, Dian and Zhao, Jeffrey and et. al.},
  booktitle={Advances in Neural Information Processing Systems},
  volume={36},
  year={2023},
}

@inproceedings{besta2024graphOfThoughts,
  title={Graph of Thoughts: Solving Elaborate Problems With Large Language Models},
  author={Besta, Maciej and Blach, Nils and Kubicek, Ales and et. al.},
  booktitle={Proceedings of the AAAI Conference on Artificial Intelligence},
  volume={38},
  number={16},
  pages={17682--17690},
  year={2024},
}

@INBOOK{6577992,
  author={Agarwal, Priyanshu and Kumar, Suren and Ryde, Julian and et. al.},
  booktitle={Robotics: Science and Systems VIII}, 
  title={Towards A Swarm of Agile Micro Quadrotors}, 
  year={2013},
  volume={},
  number={},
  pages={217-224},
  publisher={IEEE},
  doi={}
}

@INPROCEEDINGS{1242572,
  author={Alighanbari, M. and Kuwata, Y. and How, J.P.},
  booktitle={Proceedings of the 2003 American Control Conference, 2003.}, 
  title={Coordination and control of multiple UAVs with timing constraints and loitering}, 
  year={2003},
  volume={6},
  number={},
  pages={5311-5316 vol.6},
  doi={10.1109/ACC.2003.1242572}}

@inbook{doi:10.2514/6.2015-0599,
author = {Daniel Morgan and Soon-Jo Chung and Fred Y. Hadaegh},
title = {Swarm Assignment and Trajectory Optimization Using Variable-Swarm, Distributed Auction Assignment and Model Predictive Control},
booktitle = {AIAA Guidance, Navigation, and Control Conference},
chapter = {},
pages = {},
doi = {10.2514/6.2015-0599},
URL = {https://arc.aiaa.org/doi/abs/10.2514/6.2015-0599},
eprint = {https://arc.aiaa.org/doi/pdf/10.2514/6.2015-0599}
}

@INPROCEEDINGS{zhou2021ego,
  author={Zhou, Xin and Zhu, Jiangchao and Zhou, Hongyu and et. al.},
  booktitle={2021 IEEE International Conference on Robotics and Automation (ICRA)}, 
  title={EGO-Swarm: A Fully Autonomous and Decentralized Quadrotor Swarm System in Cluttered Environments}, 
  year={2021},
  volume={},
  number={},
  pages={4101-4107},
  doi={10.1109/ICRA48506.2021.9561902}}

@article{zhou2021fuel,
  title={Fuel: Fast uav exploration using incremental frontier structure and hierarchical planning},
  author={Zhou, Boyu and Zhang, Yichen and Chen, Xinyi and et. al.},
  journal={IEEE Rob. Autom. Lett.},
  volume={6},
  number={2},
  pages={779--786},
  year={2021},
  publisher={IEEE}
}

@inproceedings{feng2022fcplanner,
  title={Fc-planner: A skeleton-guided planning framework for fast aerial coverage of complex 3d scenes},
  author={Feng, Chen and Li, Haojia and Zhang, Mingjie and et. al.},
  booktitle={2024 IEEE International Conference on Robotics and Automation (ICRA)},
  pages={8686--8692},
  year={2024},
  organization={IEEE}
}

@inproceedings{li2025airswarm,
  title={Airswarm: Enabling cost-effective multi-uav research with cots drones},
  author={Li, Xiaowei and Xu, Kuan and Liu, Fen and et. al.},
  booktitle={2025 IEEE/RSJ International Conference on Intelligent Robots and Systems (IROS)},
  pages={21487--21494},
  year={2025},
  organization={IEEE}
}

@article{kim2025raven,
  title={RAVEN: Resilient Aerial Navigation via Open-Set Semantic Memory and Behavior Adaptation},
  author={Kim, Seungchan and Alama, Omar and Kurdydyk, Dmytro and et. al.},
  journal={arXiv preprint arXiv:2509.23563},
  year={2025}
}

@inproceedings{xing2023autonomous,
  title={Autonomous power line inspection with drones via perception-aware MPC},
  author={Xing, Jiaxu and Cioffi, Giovanni and Hidalgo-Carri{\'o}, Javier and et. al.},
  booktitle={2023 IEEE/RSJ International Conference on Intelligent Robots and Systems (IROS)},
  pages={1086--1093},
  year={2023},
  organization={IEEE}
}

@article{zhang2022aerial,
  title={Aerial additive manufacturing with multiple autonomous robots},
  author={Zhang, Ketao and Chermprayong, Pisak and Xiao, Feng and et. al.},
  journal={Nature},
  volume={609},
  number={7928},
  pages={709--717},
  year={2022},
  publisher={Nature Publishing Group UK London}
}

@article{mcguire2019minimal,
  title={Minimal navigation solution for a swarm of tiny flying robots to explore an unknown environment},
  author={McGuire, Kimberly N and De Wagter, Christophe and Tuyls, Karl and et. al.},
  journal={Science Robotics},
  volume={4},
  number={35},
  pages={eaaw9710},
  year={2019},
  publisher={American Association for the Advancement of Science}
}

@article{petravcek2021large,
  title={Large-scale exploration of cave environments by unmanned aerial vehicles},
  author={Petr{\'a}{\v{c}}ek, Pavel and Kr{\'a}tk{\`y}, V{\'\i}t and Petrl{\'\i}k, Mat{\v{e}}j and et. al.},
  journal={IEEE Rob. Autom. Lett.},
  volume={6},
  number={4},
  pages={7596--7603},
  year={2021},
  publisher={IEEE}
}



\end{document}